%% file: strategic-decisions-examples.tex
\documentclass[a4paper, 10pt]{article}

\usepackage{authblk}
\usepackage{booktabs} 
\usepackage[ruled]{algorithm2e} 
\usepackage[semicolon]{natbib}
\usepackage{fullpage}

\SetAlFnt{\small}
\SetAlCapFnt{\small}
\SetAlCapNameFnt{\small}
\SetAlCapHSkip{0pt}
\IncMargin{-\parindent}

\usepackage{subfig}
\usepackage{tikz}
\usepackage{Definitions}
\usepackage{algorithmic}
\newcommand{\given}{{\,|\,}}
\newcommand{\xhdr}[1]{\vspace{1mm} \noindent{{\bf #1.}}}

\definecolor{mypink}{RGB}{242,88,118}
\definecolor{mygreen}{RGB}{67,165,38}

\graphicspath{{./FIG/}}

\begin{document}

\title{Decisions, Counterfactual Explanations and Strategic Behavior}
\author{Stratis Tsirtsis}
\author{Manuel Gomez-Rodriguez}
\affil{Max Planck Institute for Software Systems \\ \{stsirtsis, manuelgr\}@mpi-sws.org}

\date{}

\maketitle

\begin{abstract}
    \input{000abstract}
\end{abstract}

\section{Introduction}
\label{sec:introduction}
\input{010introduction}
\section{Problem Formulation}
\label{sec:preliminaries}
\input{020preliminaries}

\section{Finding the optimal counterfactual explanations for a policy}
\label{sec:fixed}
\input{030fixed}

\section{Finding the optimal policy and counterfactual explanations}
\label{sec:codesign}
\input{040codesign}

\section{Increasing the diversity of the counterfactual explanations}
\label{sec:matroids}
\input{050matroids}

\section{Experiments}
\label{sec:experiments}
\input{060experiments}

\section{Conclusions}
\label{sec:conclusions}
\input{070conclusions}

\bibliographystyle{plainnat}
\bibliography{strategic-decisions-examples}

\clearpage
\newpage
\appendix

\label{sec:appendix}
\input{080appendix}
\end{document}

%% file: 000abstract.tex
%
As data-driven predictive models are increasingly used to inform decisions, 
it has been argued that decision makers should provide explanations that help individuals understand what would have to change for 
these decisions to be beneficial ones.
%
%
However, there has been little discussion on the possibility that individuals may use the above \emph{counterfactual explanations} 
to invest effort strategically and maximize their chances of receiving a beneficial decision.
In this paper, our goal is to find po\-li\-cies and counterfactual explanations that are optimal in terms of utility in such a strategic setting.
We first show that, given a pre-defined policy, the problem of finding the optimal set of counterfactual explanations is 
NP-hard.
Then, we show that the corresponding objective is nondecreasing and satisfies submodularity and this allows a standard greedy 
algorithm to enjoy approximation guarantees. 
In addition, we further show that the problem of jointly finding both the optimal po\-li\-cy and set of counterfactual explanations reduces to 
maximizing a non-monotone submodular function. As a result, we can use a recent randomized algorithm to solve the problem, 
which also offers approximation guarantees. 
Finally, we demonstrate that, by incorporating a matroid constraint into the problem formulation, we can increase the diversity of the 
optimal set of counterfactual explanations and incentivize individuals across the whole spectrum of the population to self improve.
Experiments on synthetic and real lending and credit card data illustrate our theoretical findings and show that the counterfactual 
explanations and decision policies found by our algorithms achieve higher utility than several competitive baselines.

%
%
%
%
%

%% file: 010introduction.tex
%

Whenever a bank decides to offer a loan to a customer, a university decides to admit a prospective student, or a company decides to hire a new employee, the
decision is increasingly informed by a data-driven predictive model.
In all these high-stakes applications, the goal of the predictive model is to provide accurate predictions of the outcomes from a set of observable features while
the goal of the decision maker is to take decisions that ma\-xi\-mi\-ze a given utility function.
%
%
For example, in university admissions, the predictive model may estimate the ability of each prospective student to successfully complete the graduate 
program while the decision maker may weigh the model'{}s estimate against other socio-economic considerations (\eg, number of available scholarships, 
diversity commitments).
%

In this context, there has been a tremendous excitement on the potential of data-driven predictive models to enhance decision making in high-stakes applications.
However, there has also been a heated debate about their lack of transparency and explainability~\citep{doshi2017towards,weller2017challenges,lipton2018mythos,gunning2019darpa, rudin2019stop}.
As a result, there already exists a legal requirement to grant individuals who are subject to (semi)-automated decision making the \emph{right-to-explanation} in the 
European Union~\citep{voigt2017eu, wachter2017right}.
With this motivation, there has been a flurry of work on interpretable machine learning~\citep{ribeiro2016should, koh2017understanding, lundberg2017unified, chakraborty2017interpretability, wachter2017counterfactual, murdoch2019definitions, ustun2019actionable, karimi2019model, mothilal2019explaining}, 
which has predominantly focused on developing methods to find explanations for the predictions made by a predictive model.
Within this line of work, the work most closely related to ours~\citep{wachter2017counterfactual, ustun2019actionable, karimi2019model, mothilal2019explaining} aims to find counterfactual 
explanations that help individuals understand what would have to change for a predictive model to make a positive prediction about them.
However, none of these works distinguish between decisions and predictions and, consequently, cannot be readily used to provide explanations to the decisions taken by
a decision maker, which are ultimately what individuals who are subject to (semi)-automated decision making typically care about.

In our work, we build upon a recent line of work that explicitly distinguishes between predictions and decisions~\citep{corbett2017algorithmic, kilbertus2019fair, kleinberg2018human, mitchell2018prediction, tabibian2020optimal, valera2018enhancing} and then pursue the development of methods to find counterfactual explanations for the decisions taken by a decision 
maker who is assisted by a data-driven predictive model. These counterfactual explanations help individuals understand what would have to change in order to receive a beneficial 
decision, rather than a positive prediction.  
%
%
%
%
%
%
%
Moreover, once we focus on explaining decisions, we cannot overlook the possibility that individuals may use these explanations to invest effort strategically in order 
to maximize their chances of receiving a beneficial decision.
However, this is also an opportunity for us to find counterfactual explanations that help individuals to self-improve and eventually increase the utility of a 
decision policy, as noted by several studies in economics~\citep{coate1993will, fryer2013valuing, hu2018short} and, more recently, in the computer science 
literature~\citep{kleinberg2019classifiers, perdomo2020performative, tabibian2020optimal}.
For example, if a bank explains to a customer that, if she reduces her credit card debt by 20\%, she will receive the loan she is applying for, she may feel 
compelled to reduce her overall credit card debt by the proposed percentage to pay less interest, improving her financial situation, and this will eventually 
increase the profit the bank makes when she is able to successfully return the loan.
This is in contrast with previous work on interpretable machine learning, which have ignored the influence that (counterfactual) explanations (of predictions 
by a predictive model) may have on the accuracy of predictive models and the utility of the decision policies\footnote{Refer to Appendix~\ref{app:further-related-work} for 
a discussion of further related work.}.

\xhdr{Our contributions} We cast the above problem as a Stackelberg game in which the decision maker moves first and shares her counterfactual explanations before 
individuals best-respond to these explanations and invest effort to receive a beneficial decision.
In this context, we assume that the decision maker takes decisions based on low dimensional feature vectors since, in many realistic scenarios, the 
data is summarized by just a small number of summary statistics (\eg, FICO scores)~\citep{hardt2016equality,liu2018delayed}.
Under this problem formulation, we first show that, given a pre-defined policy, the problem of finding the optimal set of counterfactual explanations is NP-hard by 
using a novel reduction of the Set Cover problem~\citep{karp1972reducibility}.
Then, we show that the corresponding objective function is monotone and submodular and, as a direct consequence, it readily follows that a standard greedy 
algorithm offers approximation guarantees.
In addition, we show that, given a pre-defined set of counterfactual explanations, the optimal policy is deterministic and can be computed in polynomial time. 
Moreover, building on this result, we can reduce the problem of jointly finding both the optimal policy and set of counterfactual explanations to maximizing a non-monotone 
submodular function.
As a consequence, we can use a recent randomized algorithm to solve the problem, which also offers approximation guarantees. 
Further, we demonstrate that, by incorporating a matroid constraint into the problem formulation, we can increase the diversity of the optimal set of counterfactual
explanations and incentivize individuals across the whole spectrum of the population to self improve.
Experiments using real lending and credit card data illustrate our theoretical findings and show that the counterfactual explanations and decision policies found by the 
above algorithms achieve higher utility than several competitive baselines\footnote{An open-source implementation can be found at \href{https://github.com/Networks-Learning/strategic-decisions}{https://github.com/Networks-Learning/strategic-decisions}.}.

%% file: 020preliminaries.tex
Given an individual with a feature vector $\xb \in \{1,...,n\}^d$ and a (\emph{ground-truth}) label $y \in \{0, 1\}$,
we assume a decision $d(\xb)\in\{0,1\}$ controls whether the corresponding label is \emph{realized}\footnote{Without 
loss of generality, we assume each feature takes $n$ different values.}. 
%
%
%
%
This setting fits a variety of real-world scenarios, where continuous features are often discretized into (percentile) 
ranges.
For example, in university admissions, the decision specifies whether a student is admitted ($d(\xb)=1$) or rejected 
($d(\xb)=0$);
the label indicates whether the student completes the program ($y=1$) or drops out ($y=0$) upon acceptance;
and the feature vector ($\xb$) may include her GRE scores, undergraduate GPA percentile, or research experience. 
%
%
Throughout the paper, we will denote the set of feature values as $\Xcal=\{\xb_1,\xb_2,\ldots,\xb_m\}$, where $m=n^d$ denotes the number of feature
values, and assume that the number of features $d$ is small, as discussed previously.

Each decision is sampled from a decision policy
$d(\xb)\sim\pi(d\given\xb)$, where, for brevity, we will write $\pi(\xb)=\pi(d=1\given\xb)$.
For each individual, the label $y$ is sampled from a conditional probability distribution $y \sim P(y\given\xb)$ and, without loss of generality, we index the 
feature values in decreasing order with respect to their corresponding outcome, \ie, $i < j \Rightarrow P(y = 1 \given \xb_i) \geq P(y = 1 \given \xb_j)$.
%
%
%
Moreover, we adopt a Stackelberg game-theoretic formulation in which each individual with initial feature value $\xb_i$ receives a (counterfactual) explanation 
from the decision maker by means of a feature value $\Ecal(\xb_i) \in \Acal \subseteq \Pcal_\pi := \{\xb\in\Xcal : \pi(\xb)=1\}$ before she (best-)responds\footnote{In 
practice, individuals with initial feature values $\xb_i$ such that $\pi(\xb) = 1$ may not receive any explanation since they are guaranteed to receive a positive decision.}.
This formulation fits a variety of real-world applications. 
For example, insurance companies often provide online car insurance simulators that, on the basis of a customer'{}s initial feature value $\xb_i$, let the customer know 
whether they are eligible for a particular deal. In case the customer does not qualify, the simulator could provide a counterfactual example $\Ecal(\xb_i)$ under which 
the individual is guaranteed to be eligible.
%
%
In the remainder, we will refer to $\Acal$ as the set of counterfactual explanations and, for each individual with initial feature value $\xb_i$, we will assume she does
not know anything about the other counterfactual explanations $\Acal \backslash \Ecal(\xb_i)$ other individuals may receive nor the decision policy $\pi(\xb)$.

Now, let $c(\xb, \Ecal(\xb_i))$ be the cost\footnote{In practice, the cost for each pair of feature values may be given by a parameterized function.} an individual pays for 
changing from $\xb_i$ to $\Ecal(\xb_i)$ and $b(\pi, \xb) = \EE_{d \sim \pi(d \given x)}[d(\xb)]$ be the (immediate) benefit she obtains from a policy $\pi$, which is just the 
probability that the individual receives a positive decision.
%
Then, following~\citet{tabibian2020optimal}, each individual'{}s best response is to change from her initial feature value $\xb_i$ to 
$\Ecal(\xb_i)$ iff the gained benefit she would obtain outweighs the cost she would pay for changing features, \ie,
\begin{equation*}
\Ecal(\xb_i) \in \{ \xb_j\in\Xcal \,:\, b(\pi, \xb_j) - c(\xb_i, \xb_j) \geq b(\pi, \xb_i) \} := \Rcal(\xb_i),
\end{equation*}
and it is to keep her initial feature value $\xb_i$ otherwise. Here, we will refer to $\Rcal(\xb_i)$ as the \emph{region of adaptation}.
%
Then, at a population level, the above best response results into a transportation of mass between the original feature distribution $P(\xb)$ and a new feature distribution 
$P(\xb \given \pi, \Acal)$ induced by the policy $\pi$ and the counterfactual explanations $\Acal$. More specifically, we can readily derive an analytical expression for 
the induced feature distribution in terms of the original feature distribution, \ie, for all $\xb_j \in \Xcal$,
\begin{equation*}
P(\xb_j \given \pi, \Acal) = P(\xb_j)\II(\Rcal(\xb_j)\cap\Acal=\emptyset) + \sum_{i \in [m]} P(\xb_i) \II(\Ecal(\xb_i) = \xb_j \wedge \xb_j\in\Rcal(\xb_i)), 
\end{equation*}
Similarly as in previous work~\citep{corbett2017algorithmic, valera2018enhancing, kilbertus2019fair, tabibian2020optimal}, we will assume that the decision maker 
is rational, has access to (an estimation of) the original feature distribution $P(\xb)$, and aims to maximize the (immediate) utility $u(\pi, \gamma)$, which is the 
expected overall profit she obtains, \ie,
\begin{equation} \label{eq:utility}
\begin{split}
u(\pi, \Acal) & = \EE_{\xb \sim P(\xb \given \pi, \Acal), y \sim P(y \given \xb), d \sim \pi(\xb)} \left[ y d(\xb) - \gamma d(\xb) \right] \\
& = \EE_{\xb \sim P(\xb \given \pi, \Acal)} \left[ \pi(\xb) (P(y = 1 \given \xb) - \gamma) \right], 
\end{split}
\end{equation}
where $\gamma \in (0, 1)$ is a given constant reflecting economic considerations of the decision maker. For example, in university admissions, the term
$\pi(\xb) P(y = 1 \given \xb)$ is proportional to the expected number of students who are admitted and complete the program, the term $\pi(\xb) \gamma$
is proportional to the number of students who are admitted, and $\gamma$ measures the cost of education in units of graduated students.
As a direct consequence, given a feature value $\xb_i$ and a set of counterfactual explanations $\Acal$, we can conclude that, if $\Rcal(\xb_i) \cap \Acal \neq \emptyset$, 
the decision maker will decide to provide the counterfactual explanation $\Ecal(\xb_i)$ that provides the largest utility gain under the assumption that individuals best 
respond, \ie, 
\begin{equation} \label{eq:best-counterfactual-explanations-rational}
\Ecal(\xb_i) = \argmax_{\xb \in \Acal \cap \Rcal(\xb_i)} P(y \given \xb) \,\, \text{for all} \,\, \xb_i \in \Xcal\setminus\Pcal_\pi \,\, \text{such that} \,\, \Rcal(\xb_i) \cap \Acal \neq \emptyset,
\end{equation}
and, if $\Rcal(\xb_i) \cap \Acal = \emptyset$, we arbitrarily assume that $\Ecal(\xb_i) = \argmin_{\xb \in \Acal} c(\xb_i, \xb)$\footnote{Note that, if $\Acal \cap \Rcal(\xb_i) = \emptyset$,
the individual'{}s best response is to keep her initial feature value $\xb_i$ and thus any choice of counterfactual explanation $\Ecal(\xb_i)$ leads to the same utility.}. 

Given the above preliminaries, our goal is to help the decision maker to first find the optimal set of counterfactual explanations $\Acal$ for a pre-defined
policy in Section~\ref{sec:fixed} and then both the optimal policy $\pi$ and set of counterfactual explanations $\Acal$ in Section~\ref{sec:codesign}. 
%
%
%

%
\xhdr{Remarks}
Given an individual with initial feature value $\xb$, one may think that, by providing the counterfactual explanation $\Ecal(\xb) \in \Acal \cap \Rcal(\xb)$ that gives the largest 
utility gain, the decision maker is not acting in the individual'{}s best interest but rather selfishly. 
This is because there may exist another counterfactual explanation $\Ecal_m(\xb) \in \Acal \cap \Rcal(\xb)$ with lower cost for the individual, \ie, $c(\xb, \Ecal_m(\xb)) \leq c(\xb, \Ecal(\xb))$.
In our work, we argue that the provided counterfactual explanations help the individual to achieve a greater self-improvement and this is likely to result in a superior long-term 
well-being, as illustrated in Figure~\ref{fig:synth}(c) in Appendix~\ref{app:synthetic}.
For example, consider a bank issuing credit cards who wants to maintain credit for trustworthy customers and incentivize the more risky ones to improve their financial 
status.
In this case, $\Ecal(\xb)$ is the explanation that maximally improves the financial status of the individual, making the repayment more likely, but requires her to pay a larger 
(immediate) cost. 
In contrast, $\Ecal_m(\xb)$ is an alternate explanation that requires the individual to pay a smaller (immediate) cost but, in comparison with $\Ecal(\xb)$, would result in a 
higher risk of default. 
In this context, note that the individual would be ``willing'' to pay the cost of following either $\Ecal(\xb)$ or $\Ecal_m(\xb)$ since both explanations lie within the region of adaptation 
$\Rcal(\xb)$. %
We refer the interested reader to Appendix~\ref{app:realistic} for an anecdotal real-world example of $\Ecal(\xb)$ and $\Ecal_m(\xb)$.

As argued very recently~\citep{kleinberg2019classifiers,miller2019strategic, tabibian2020optimal}, due to Goodhart'{}s law, the conditional probability $P(y \given \xb)$ may change after individuals 
(best)-respond if the true causal effect between the observed features $\xb$ and the outcome variable $y$ is partially described by unobserved features.
Moreover,~\citet{miller2019strategic} have argued that, to distinguish between gaming and improvement, it is necessary to have access to the full underlying causal graph between
the features and the outcome variable.
In this work, for simplicity, we assume that $P(y \given \xb)$ does not change, however, it would be very interesting to lift this assumption in future work.

%% file: 030fixed.tex
In this section, our goal is to find the optimal set of counterfactual explanations $\Acal^{*}$ for a pre-defined policy $\pi$, \ie,
\begin{equation} \label{eq:optimal-counterfactual-explanations}
\Acal^* = \argmax_{\Acal \subseteq \Pcal_\pi \,: \, |\Acal| \leq k} u(\pi,\Acal),
\end{equation} 
where the cardinality constraint on the set of counterfactual explanations balances the decision maker'{}s obligation to 
be transparent with trade secrets~\citep{barocas2020hidden}. More specifically, note that, without this constraint, an 
adversary could reverse-engineer the entire decision policy $\pi(\xb)$ by impersonating individuals with different feature 
values $\xb$~\citep{css}.
%
%

%
As it will become clearer in the experimental evaluation in Section~\ref{sec:experiments}, our results may persuade decision makers to be 
transparent about their decision policies, something they are typically reluctant to be despite the increasing legal requirements, since we 
show that transparency increases the utility of the policies.
%
%
%
%
%
%
Moreover, throughout this section, we will assume that the decision maker who picks the pre-defined policy is rational\footnote{Note that, if the decision maker is 
rational and her goal is to maximize the utility, as defined in Eq.~\ref{eq:utility}, then, for all $\xb \in \Xcal$ such that $P(y=1\given\xb)<\gamma$, it holds that $\pi(\xb)=0$.} 
and the policy is outcome monotonic\footnote{A policy $\pi$ is called outcome monotonic if $P(y=1\given\xb_i)\geq P(y=1\given\xb_j) \Leftrightarrow \pi(\xb_i)\geq\pi(\xb_j)\  \forall \xb_i,\xb_j\in\Xcal$.}\footnote{If the policy $\pi$ is deterministic, our results also hold for non outcome monotonic policies.}~\citep{tabibian2020optimal}. Outcome 
monotonicity just implies that, the higher an individual'{}s outcome $P(y = 1 \given \xb)$, the higher their chances of receiving a positive decision $\pi(\xb)$.
%
%
%
%

%
%
%
%

%

%
%
Unfortunately, using a novel reduction of the Set Cover problem~\citep{karp1972reducibility}, the following theorem reveals that we cannot expect to find the optimal 
set of counterfactual explanations in polynomial time (proven in Appendix~\ref{app:np-hard}):
\begin{theorem}\label{theor:np-hard}
The problem of finding the optimal set of counterfactual explanations that maximizes utility under a cardinality constraint is NP-Hard.
\end{theorem}

Even though Theorem~\ref{theor:np-hard} is a negative result, we will now show that the objective function in Eq.~\ref{eq:optimal-counterfactual-explanations} satisfies a set of desirable properties, \ie, non-negativity, monotonicity and submodularity\footnote{A function $f:2^\Xcal \rightarrow \RR$ is submodular if for every $\Acal,\Bcal\subseteq \Xcal : \Acal\subseteq \Bcal$ and $x\in\Xcal\setminus \Bcal$ it holds that $f(\Acal\cup \{x\})-f(\Acal)\geq f(\Bcal \cup \{x\})-f(\Bcal)$.}, which allow a standard greedy algorithm to enjoy approximation guarantees at solving the problem.
To this aim, with a slight abuse of notation, we first express the objective function as a set function $f(\Acal)=u(\pi,\Acal)$, which takes values over the ground
set of counterfactual explanations, $\Pcal_\pi$.
%
Then, we have the following proposition (proven in Appendix~\ref{app:monsub}):
\begin{proposition}\label{prop:monsub}
The function $f$ is non-negative, submodular and monotone.
\end{proposition}
The above result directly implies that the standard greedy algorithm~\citep{nemhauser1978analysis} for maximizing a non-negative, submodular and monotone function will find a solution $\Acal$ to the problem such that $f(\Acal) \geq (1-1/e) f(\Acal^{*})$, where $\Acal^{*}$ is the optimal set of counterfactual explanations.
The algorithm starts from a solution set $\Acal=\emptyset$ and it iteratively adds to $\Acal$ the counterfactual explanation $\xb\in\Pcal_\pi\setminus \Acal$ that provides the 
maximum marginal difference $f(\Acal\cup\{\xb\})-f(\Acal)$.
Algorithm~\ref{alg:greedy} in Appendix~\ref{app:algorithms} provides a pseudocode implementation of the algorithm.

%
%
%
Finally, since the greedy algorithm computes the marginal difference of $f$ for at most $m$ elements per iteration and, following from the proof of Proposition~\ref{prop:monsub}, the marginal difference $f(\Acal \cup \{\xb\})-f(\Acal)$ can be computed in $\Ocal(m)$, 
then it immediately follows that, in our problem, the greedy algorithm has an overall complexity of $\Ocal(km^2)$.


%% file: 040codesign.tex
In this section, our goal is to jointly find the optimal decision policy and set of counterfactual explanations $\Acal^{*}$, \ie,
\begin{equation} \label{eq:optimal-policy-and-counterfactual-explanations}
\pi^*, \Acal^* = \argmax_{(\pi,\Acal): \Acal\subseteq\Pcal_\pi \wedge |\Acal |\leq k} u(\pi,\Acal)
\end{equation} 
where, similarly as in the previous section, $k$ is the maximum number of counterfactual explanations the decision maker is willing to 
provide to the population to balance the right to explanation with trade secrets.
%
%
%
%
By jointly optimizing both the decision policy and the counterfactual explanations, we may obtain an additional gain in terms of utility in comparison with just 
optimizing for the set of counterfactual explanations given the optimal decision policy in a non-strategic setting, as shown in Figure~\ref{fig:pushed} in 
Appendix~\ref{app:pushed}. 
Moreover, as we will show in the experimental evaluation in Section~\ref{sec:experiments}, this additional gain will be significant.

Similarly as in Section~\ref{sec:fixed}, we cannot expect to find the optimal policy and set of counterfactual explanations in polynomial time. More specifically, we have the
following negative result, which easily follows from Proposition~\ref{prop:uniq} and slightly extending the proof of Theorem~\ref{theor:np-hard}:
\begin{theorem}\label{theor:np-hard2}
The problem of jointly finding both the optimal policy and the set of counterfactual explanations that maximize utility under a cardinality constraint is NP-hard.
\end{theorem}
%

However, while the problem of finding both the policy and the set of counterfactual explanations appears significantly more challenging than the 
problem of finding just the set of counterfactual explanations given a pre-defined policy (refer to Eq.~\ref{eq:optimal-counterfactual-explanations}), the following proposition 
shows that the problem is not inherently \emph{harder}. 
More specifically, for each possible set of counterfactual explanations, it shows that the policy that maximizes the utility can be easily 
computed (proven in Appendix~\ref{app:uniq}):
%
%
\begin{proposition}\label{prop:uniq}
Given a set of counterfactual explanations $\Acal \subseteq \Ycal := \{\xb\in\Xcal : P(y=1\given\xb)\geq \gamma\}$\footnote{Since the decision maker is rational, she will never provide 
an explanation that contributes negatively to her utility.}, the policy $\pi_\Acal^* = \argmax_{\pi :\Acal\subseteq\Pcal_\pi} u(\pi, \Acal)$ that maximizes the utility is deterministic and 
can be found in polynomial time, \ie, 
\begin{equation}\label{eq:expol}
\pi_\Acal^*(\xb)= 
\begin{cases}
	1 & \text{if } \xb\in\Acal \vee \{\xb' \in \Acal : P(y=1\given \xb') > P(y=1\given \xb) \wedge c(\xb,\xb')\leq 1\} = \emptyset  \wedge \xb\in\Ycal \\
    0 & \text{otherwise}.
\end{cases} 
\end{equation}
\end{proposition}
%

%
%
%
%
%
%
The above result implies that, to set all the values of the optimal decision policy, we only need to perform $\Ocal(km)$ comparisons. Moreover, it reveals that, 
in contrast with the non strategic setting, the optimal policy given a set of counterfactual explanations is not a deterministic threshold rule with a single threshold~\citep{corbett2017algorithmic, valera2018enhancing}, \ie, %
%
\begin{equation}\label{eq:dtr}
\pi(\xb)=  \begin{cases}
	1 & \text{if } P(y = 1\given\xb)\geq\gamma\\
    0 & \text{otherwise},
\end{cases} 
\end{equation}
but rather a more conservative deterministic decision policy that does not depend only on the outcome $P(y = 1 \given \xb)$ and $\gamma$ but also on the cost individuals pay to change features.
Moreover, we can build up on the above result to prove that the problem of finding the optimal decision policy and set of counterfactual explanations can be reduced to maximizing a non-monotone submodular function. 
%
%
%
%
%
To this aim, let $\pi_{\Acal}^*$ be the optimal policy induced by a given set of counterfactual explanations $\Acal$, as in Proposition~\ref{prop:uniq}, and define
the set function $h(\Acal) = u(\pi_{\Acal}^*, \Acal)$ over the ground set $\Ycal$. Then, we have the following proposition (proven in Appendix~\ref{app:nonmon}):

\begin{proposition}\label{prop:nonmon}
The function $h$ is non-negative, submodular and non-monotone.
\end{proposition}

Fortunately, there exist efficient algorithms with global approximation guarantees for maximizing a non-monotone submodular function 
under cardinality constraints.
In our work, we use the randomized polynomial time algorithm by~\citet{buchbinder2014submodular}, which can find a solution $\Acal$ such 
that $h(\Acal) \geq (1/e) h(\Acal^{*})$, where $\Acal^{*}$ and $\pi_{\Acal^{*}}^*$ are the optimal set of counterfactual explanations and decision policy, respectively.
The algorithm is just a randomized variation of the standard greedy algorithm.
It starts from a solution set $\Acal=\emptyset$ and it iteratively adds one counterfactual explanation $\xb \in \Ycal \backslash \Acal$.
However, instead of greedily choosing the element $\xb$ that provides the maximum marginal difference $h(\Acal\cup\{\xb\})-h(\Acal)$, it sorts all 
the candidate elements with respect to their marginal difference and picks one at random among the top $k$.
Algorithm~\ref{alg:randomized} in Appendix~\ref{app:algorithms} provides a pseudocode implementation of the algorithm.
%
%

Finally, since the above randomized algorithm has a complexity of $\Ocal(km)$ and, following from the proof of Proposition~\ref{prop:nonmon}, the marginal difference of $h$ can 
be computed in $\Ocal(m)$, it readily follows that, in our problem, the algorithm has a complexity of $\Ocal(km^2)$.
%
%

%
%

%% file: 050matroids.tex
In many cases, decision makers may like to ensure that individuals across the whole spectrum of the population are incentivized to self-improve.
%
%
For example, in a loan scenario, the bank may use age group as a feature to estimate the probability that a customer repays the loan, however, it may like to 
deploy a decision policy that incentivizes individuals across all age groups in order to improve the financial situation of all.
To this aim, the decision maker can increase the diversity of the optimal set of counterfactual explanations by incorporating a matroid constraint
into the problem formulation, rather than a cardinality constraint.
%

Formally, consider disjoint sets $\Xcal_1, \Xcal_2,\ldots,\Xcal_l$ such that $\bigcup_i \Xcal_i=\Xcal$ and integers $d_1,d_2,\ldots,d_l$ such that $k=\sum_i d_i$.
Then, a partition matroid is the collection of sets $\{S\subseteq 2^\Xcal: |S\cap\Xcal_i|\leq d_i\ \forall i\in[l]\}$.
In the loan example, the decision maker could search for a set of counterfactual explanations $\Acal$ within a partition matroid where each one of the $\Xcal_i$'s 
corresponds to the feature values covered by each age group and $d_i=k/l\ \forall i\in[l]$.
This way, the set of counterfactual explanations $\Acal$ would include explanations for every age group. 

In this case, the decision maker could rely on a variety of polynomial time algorithms with global guarantees for submodular function maximization under 
matroid constraints, \eg, the algorithm by~\citet{calinescu2011maximizing}.

%% file: 060experiments.tex
In this section, we evaluate Algorithms~\ref{alg:greedy} and~\ref{alg:randomized} using real loan and credit card data and show that the
counterfactual explanations and decision policies found by our algorithms achieve higher utility than several competitive baselines. 
Appendix~\ref{app:synthetic} contains additional experiments on synthetic data.

%

\xhdr{Experimental setup} 
We experiment with two publicly available datasets: (i) the \textit{lending} dataset~\citep{len}, which 
contains information about all accepted loan applications in LendingClub during the 2007-2018 period and (ii) the \textit{credit} dataset~\citep{yeh2009comparisons}, which contains information about a bank'{}s credit card payoffs\footnote{We used a version of the credit dataset preprocessed by~\citet{ustun2019actionable}}.
For each accepted loan applicant (or credit card holder), we use various demographic information and financial status indicators as features $\xb$ and the current
loan status (or credit payoff status) as label $y$.
%
%
Appendix~\ref{app:feature-representation} contains more details on the specific features we used in each dataset and also describes the procedure we followed to 
approximate $P(y \given \xb)$.
%

To set the values of the cost function $c(\xb_i, \xb_j)$, we use the maximum percentile shift among actionable features\footnote{A feature is actionable if an individual can 
change its values in order to get a positive decision.}, similarly as in~\citet{ustun2019actionable}.
More specifically, let $\Lcal$ be the set of actionable (numerical) features and $\bar{\Lcal}$ be the set of non-actionable (discrete-valued) features\footnote{In the credit dataset, 
$\bar{\Lcal}$ contains Marital Status, Age Group and Education Level and $\bar{\Lcal}$ contains the remaining features and, in the lending 
dataset, $\Lcal$ contains all features.}.
Then, for each pair of feature values $\xb_i, \xb_j$ we define the cost function as:
\begin{equation}\label{eq:cost}
c(\xb_i, \xb_j) = 
\begin{cases}
	\alpha \cdot \max_{l\in\Lcal}|Q_l(x_{j,l})-Q_l(x_{i,l})| & \text{if } x_{i,l}=x_{j,l}\ \forall l\in\bar{\Lcal} \\
    \infty & \text{otherwise},
\end{cases}
\end{equation}
where $x_{j,l}$ is the value of the $l$-th feature for the feature value $\xb_j$, $Q_l(\cdot)$ is the CDF of the numerical feature $l \in\Lcal$ 
and $\alpha\geq 1$ is a scaling factor.
As an exception, in the credit dataset, we always set the cost $c(\xb_i, \xb_j)$ between two feature values to $\infty$ if $Q_l(x_{j,l})<Q_l(x_{i,l})$ for $l\in\{\text{Total Overdue Counts}, \text{Total Months Overdue}\}$ considering the fact that history of overdue payments cannot be erased.
In this context, we would like to acknowledge that more sophisticated cost functions can be designed in terms of feasibility and difficulty of adaptation, 
taking into account domain knowledge and information about the deployed classifier, however, it goes beyond the scope of our work.

Finally, in our experiments, we compare the utility of the following decision policies and counterfactual explanations:
%

\vspace{1mm} 
\noindent --- \emph{Black box:} decisions are taken by the optimal decision policy in the non-strategic setting, given by Eq.~\ref{eq:dtr}, and individuals 
do not receive any counterfactual explanations.

\vspace{1mm}
\noindent --- \emph{Minimum cost:} decisions are taken by the optimal decision policy in the non-strategic setting, given by Eq.~\ref{eq:dtr},
and individuals receive counterfactual explanations of minimum cost with respect to their initial feature values, similarly as in previous work~\citep{ustun2019actionable, tolomei2017interpretable, karimi2019model}. 
More specifically, we cast the problem of finding the set of counterfactual explanations as the minimization of the weighted average cost individuals pay to
change their feature values to the closest counterfactual explanation, \ie, 
\begin{equation*}
\Acal_{mc} = \argmin_{\Acal \subseteq \Pcal_\pi \,: \, |\Acal| \leq k} \sum_{\xb_i\in\Xcal\setminus\Pcal_\pi} P(\xb_i) \min_{\xb_j\in\Acal}c(\xb_i,\xb_j), 
\end{equation*}
and realize that this problem is a version of the k-median problem, which we can solve using a greedy heuristic~\citep{solis2006approximation}.

\vspace{1mm} 
\noindent --- \emph{Diverse:} decisions are taken by the optimal decision policy in the non-strategic setting, given by Eq.~\ref{eq:dtr},
and individuals receive a set of diverse counterfactual explanations of minimum cost with respect to their initial feature values, similarly as in previous
work~\citep{russell2019efficient, mothilal2019explaining}, \ie,
\begin{equation*}
\Acal_{d} = \argmax_{\Acal \subseteq \Pcal_\pi \,: \, |\Acal| \leq k} \sum_{\xb_i\in\Xcal\setminus\Pcal_\pi} P(\xb_i)\II(\Rcal(\xb_i)\cap\Acal\neq\emptyset),
\end{equation*}
To solve the above problem, we realize it can be reduced to the weighted version of the maximum coverage problem, which can be solved using
a well-known greedy approximation algorithm~\citep{hochbaum1998analysis}.

\vspace{1mm} 
\noindent --- \emph{Algorithm 1:} decisions are taken by the optimal decision policy in the non-strategic setting, given by Eq.~\ref{eq:dtr}, and individuals
receive counterfactual explanations given by Eq.~\ref{eq:best-counterfactual-explanations-rational}, where $\Acal$ is found using Algorithm~\ref{alg:greedy}.

\vspace{1mm}
\noindent --- \emph{Algorithm 2:} decisions are taken by the decision policy given by Eq.~\ref{eq:expol} and individuals receive counterfactual explanations given
by Eq.~\ref{eq:best-counterfactual-explanations-rational}, where $\Acal$ is found using Algorithm~\ref{alg:randomized}.

%

%
\begin{figure}[t]
	\centering
	\subfloat[Lending dataset]{
    		 \centering
    		 \includegraphics[scale=0.22]{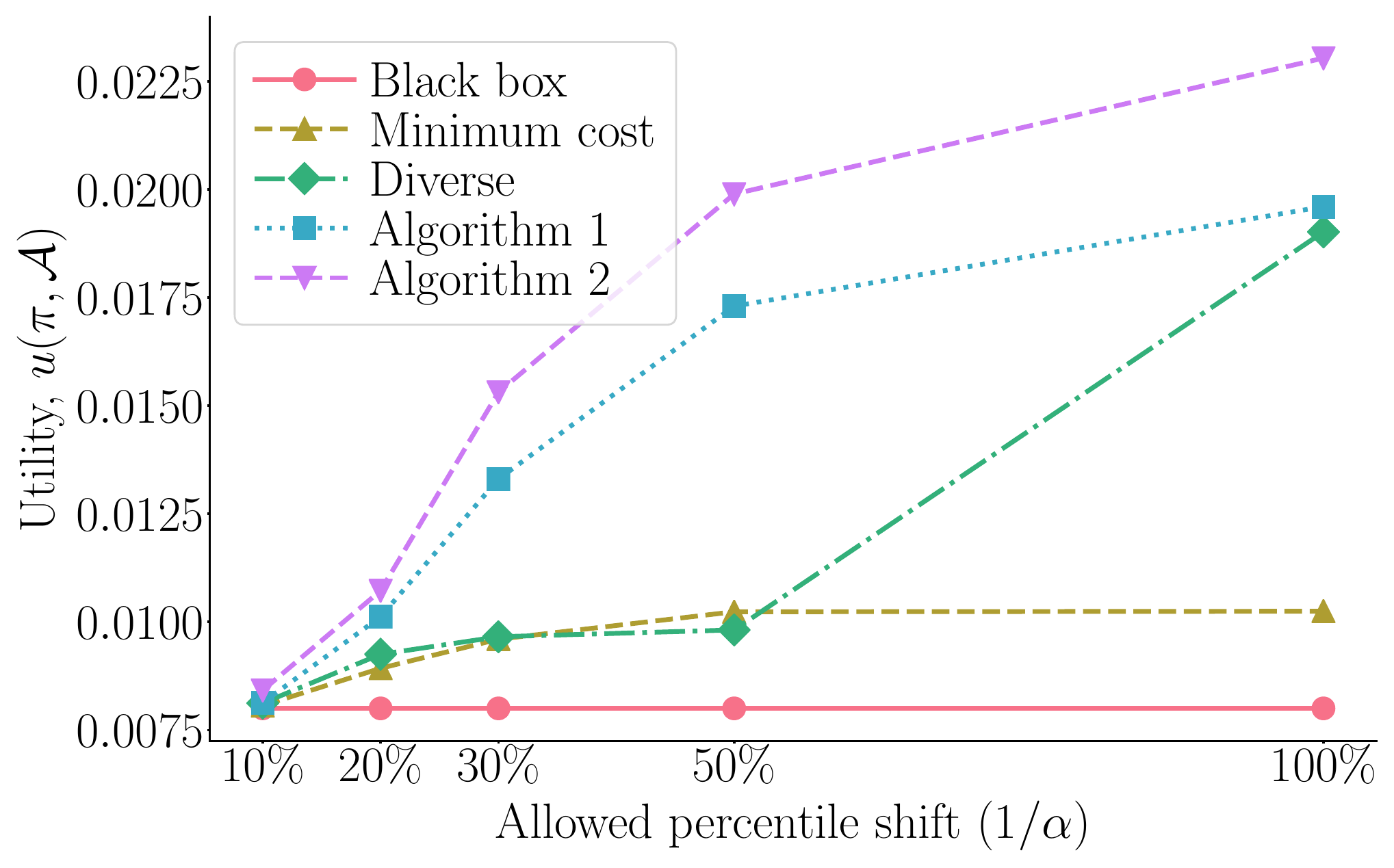}
    	}
    \hspace{10mm}
    	\subfloat[Credit dataset]{
    		 \centering
    		 \includegraphics[scale=0.22]{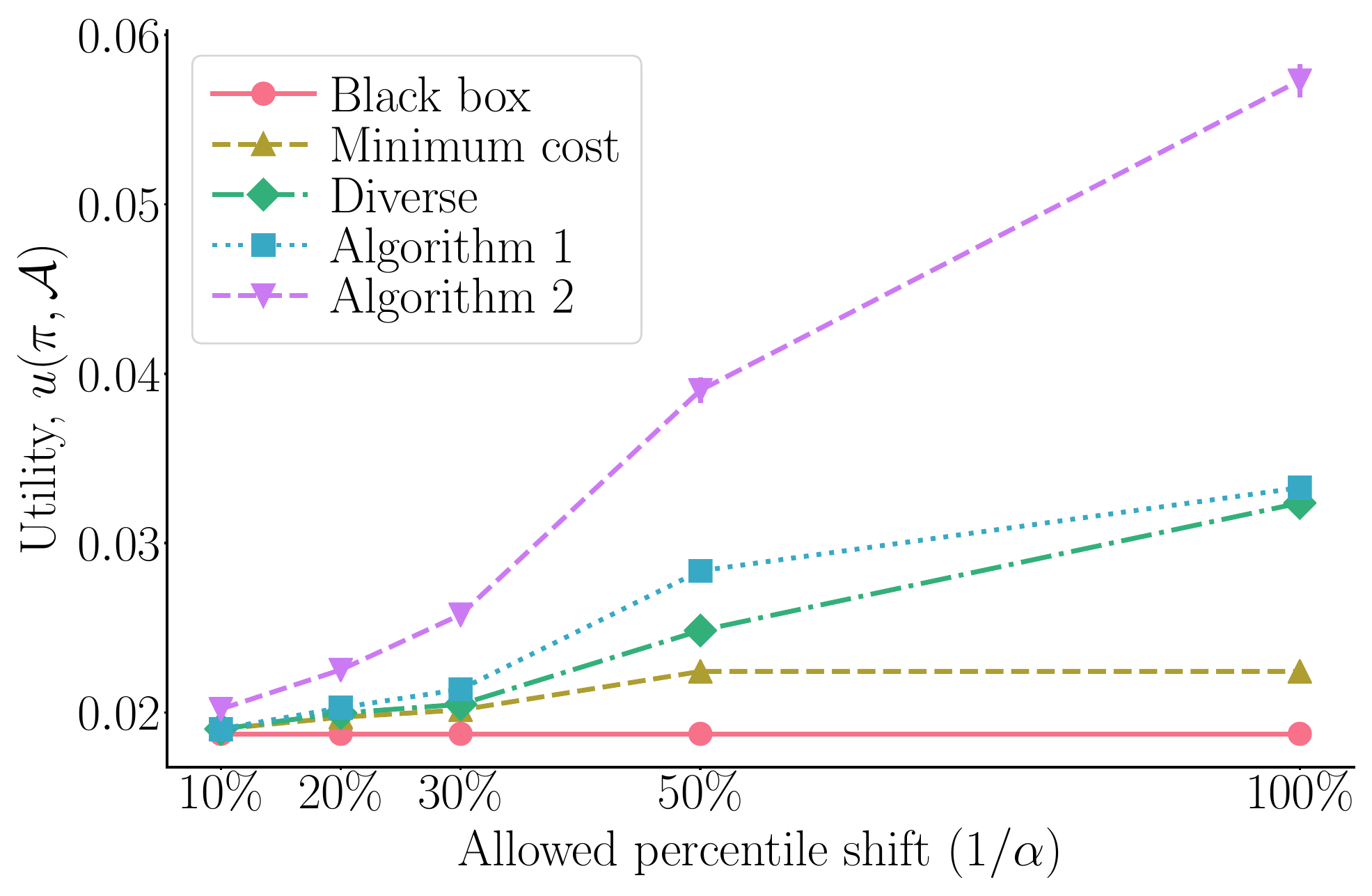}
    	}
     \caption{Utility achieved by five types of decision policies and counterfactual explanations against the value of the parameter $\alpha$, in the lending and credit datasets. 
     %
     %
     In panel (a), the number of feature values is $m=400$ and, in panel (b), it is $m=3200$. 
     In both panels, we set $k = 0.05m$ and we repeat each experiment $20$ times.}
     \label{fig:alphas}
\end{figure}

\begin{figure}[t]
	\centering
	\subfloat[Alg.~\ref{alg:greedy}, lending]{
    		 \centering
    		 \includegraphics[scale=0.20]{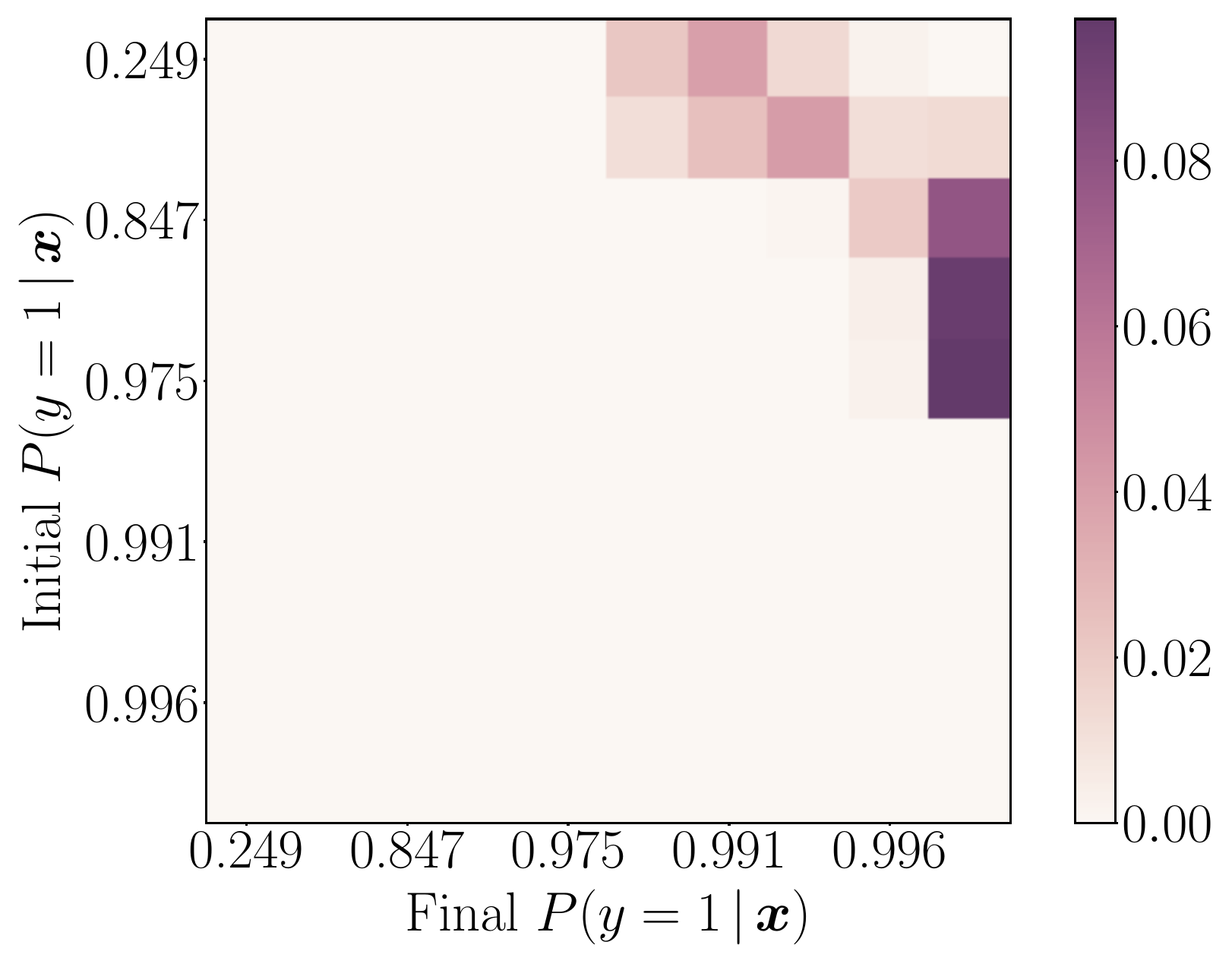}
    	}
    	\subfloat[Alg.~\ref{alg:randomized}, lending]{
    		 \centering
    		 \includegraphics[scale=0.20]{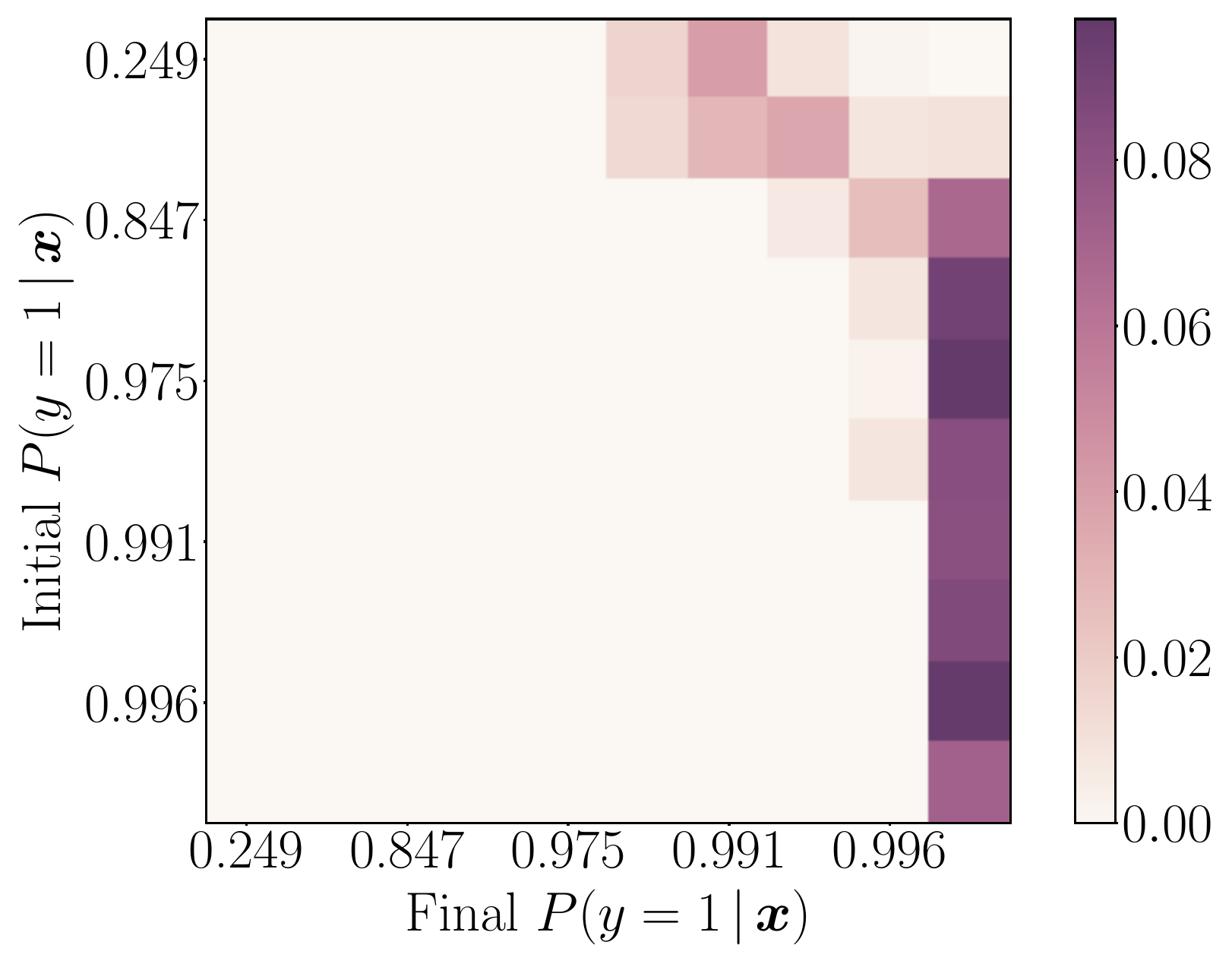}
    	}
    	\subfloat[Alg.~\ref{alg:greedy}, credit]{
    		 \centering
    		 \includegraphics[scale=0.20]{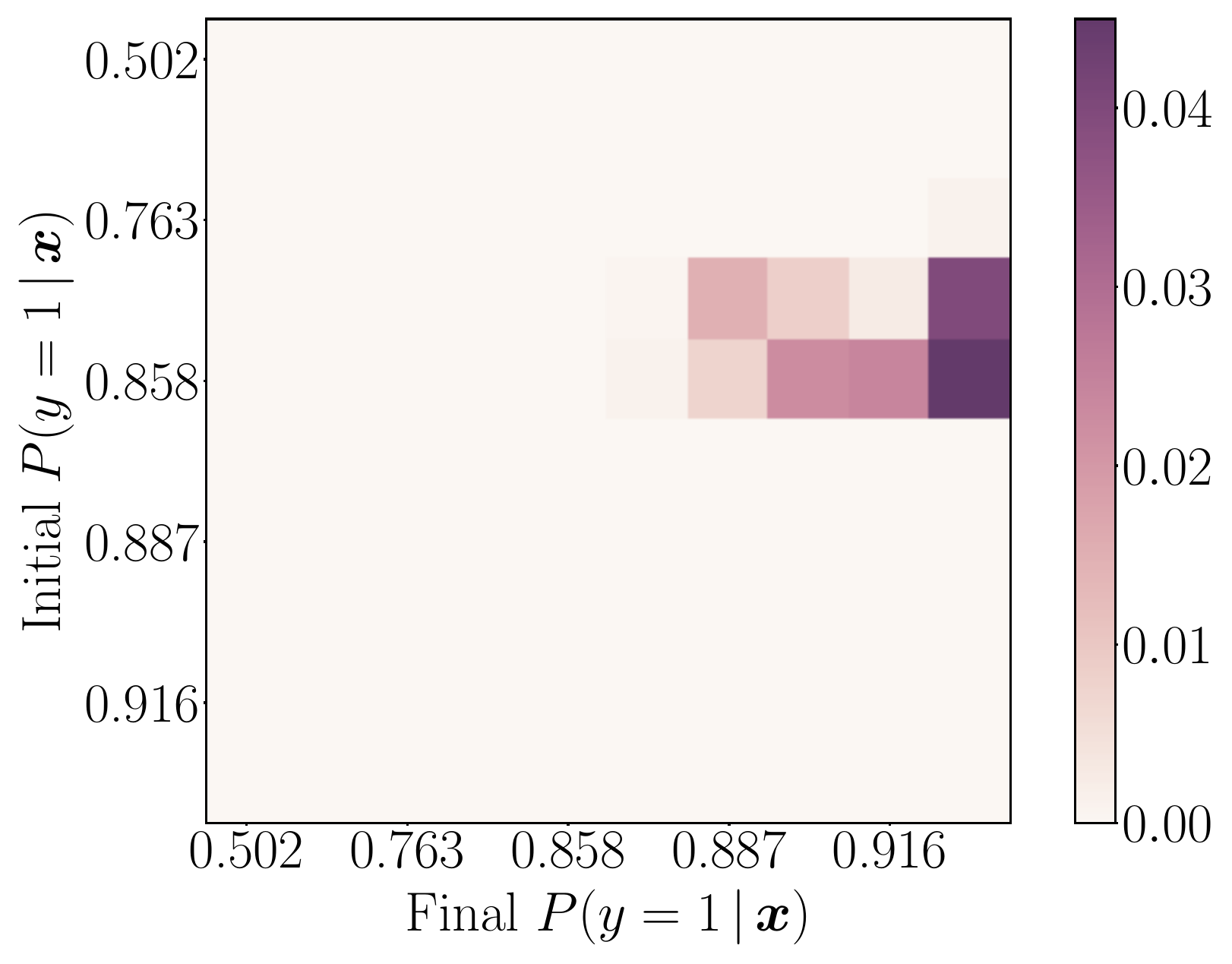}
    	}
    	\subfloat[Alg.~\ref{alg:randomized}, credit]{
    		 \centering
    		 \includegraphics[scale=0.20]{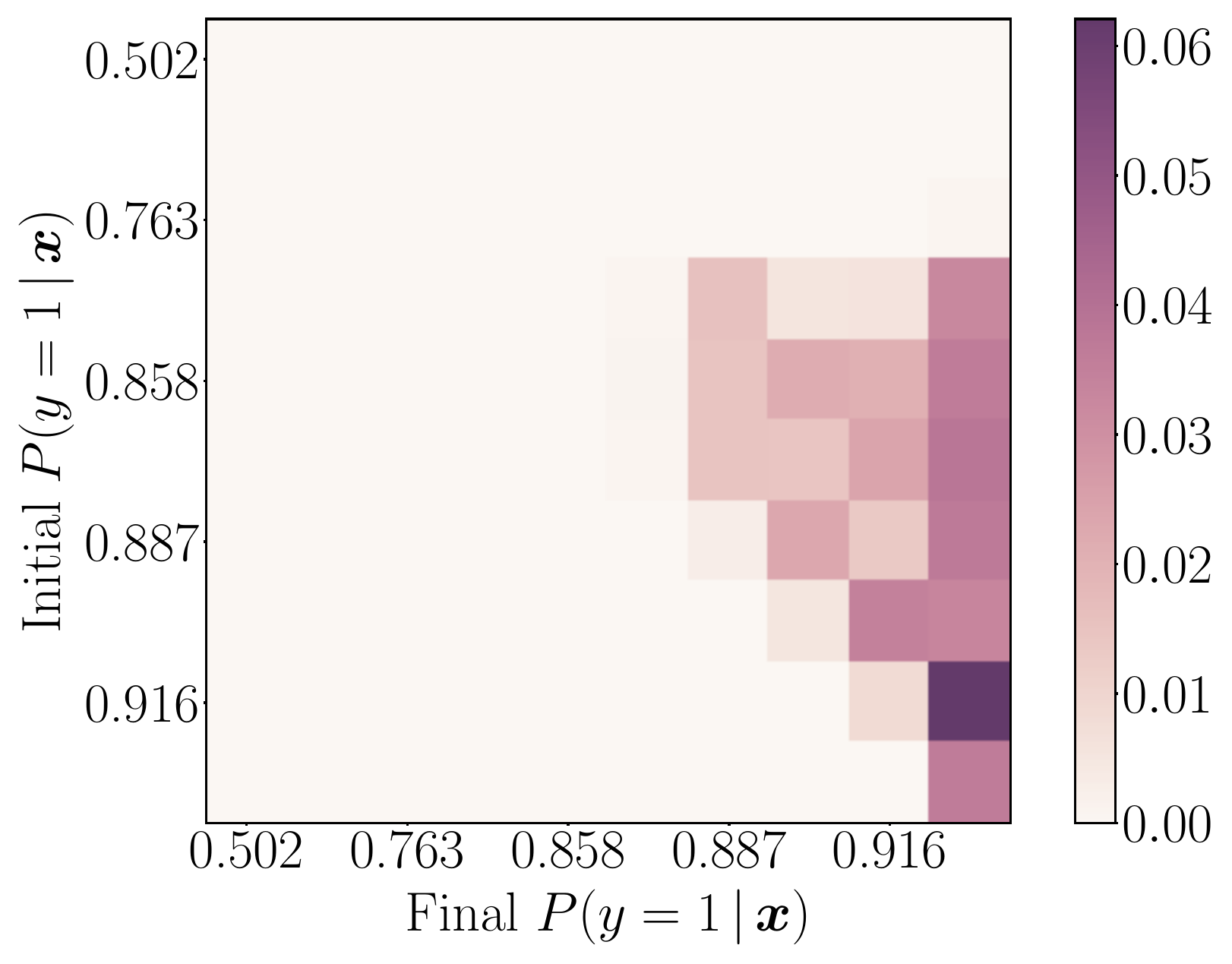}
    	}
     \caption{Transportation of mass induced by the policies and counterfactual explanations used in Algorithm~\ref{alg:greedy} and~\ref{alg:randomized} 
     in both the lending and the credit dataset. 
     For each individual in the population, whose best-response is to change her feature value, we record her outcome $P(y=1 \given \xb)$ before
     the best response (Initial $P(y=1 \given \xb)$) and after the best response (Final $P(y=1 \given \xb)$). 
     In each panel, the color is proportional to the percentage of individuals who move from initial $P(y=1 \given \xb)$ to final $P(y=1 \given \xb)$
     and we set $\alpha = 2$.}
     \label{fig:transp}
\end{figure}

\begin{figure}[t]
	\centering
    	\subfloat[Utility vs. $k$]{
    		 \centering
    		 \includegraphics[scale=0.22]{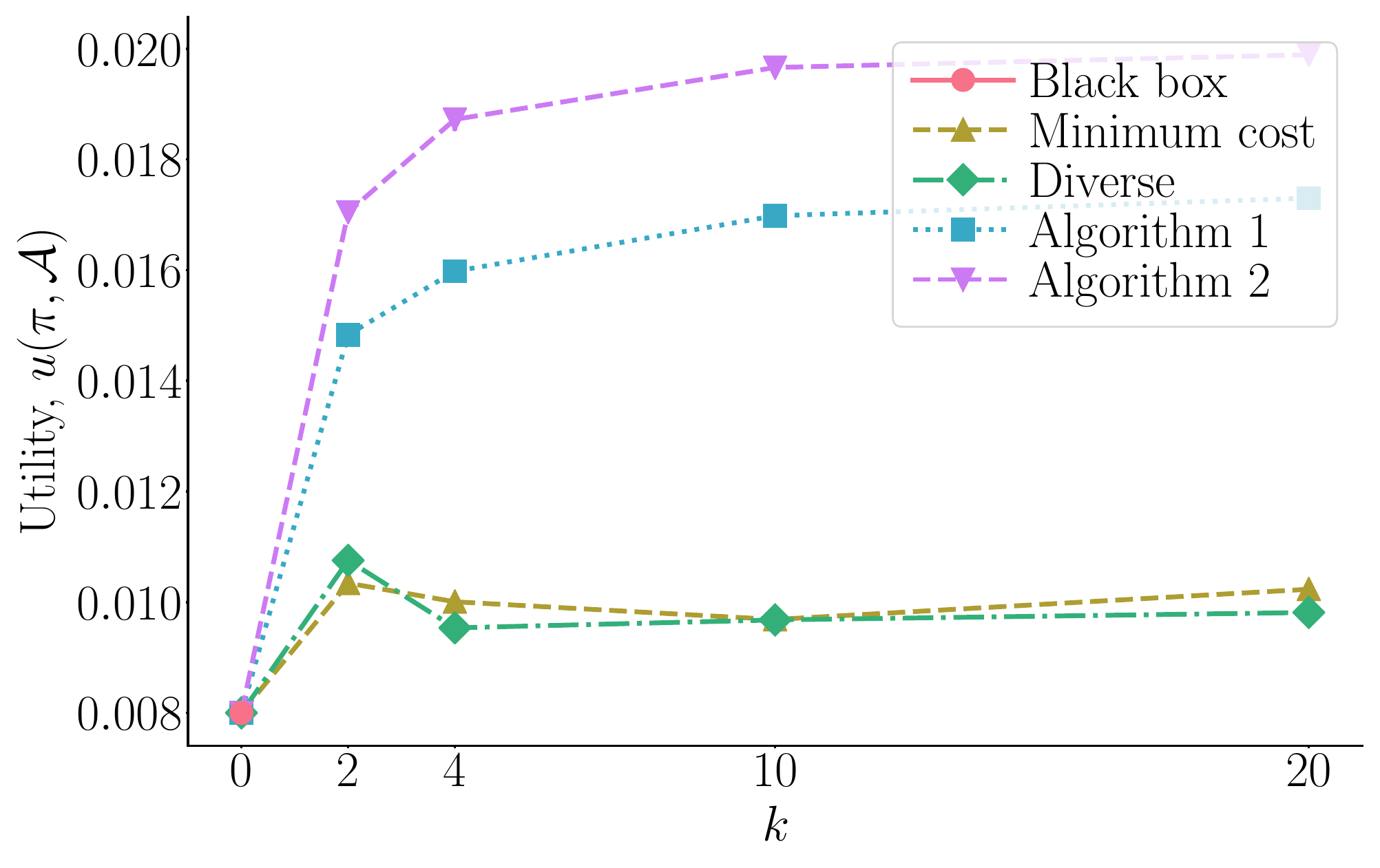}
    	}
  	\hspace{10mm}
    	\subfloat[Utility vs. $k$ under leakage]{
    		 \centering
    		 \includegraphics[scale=0.22]{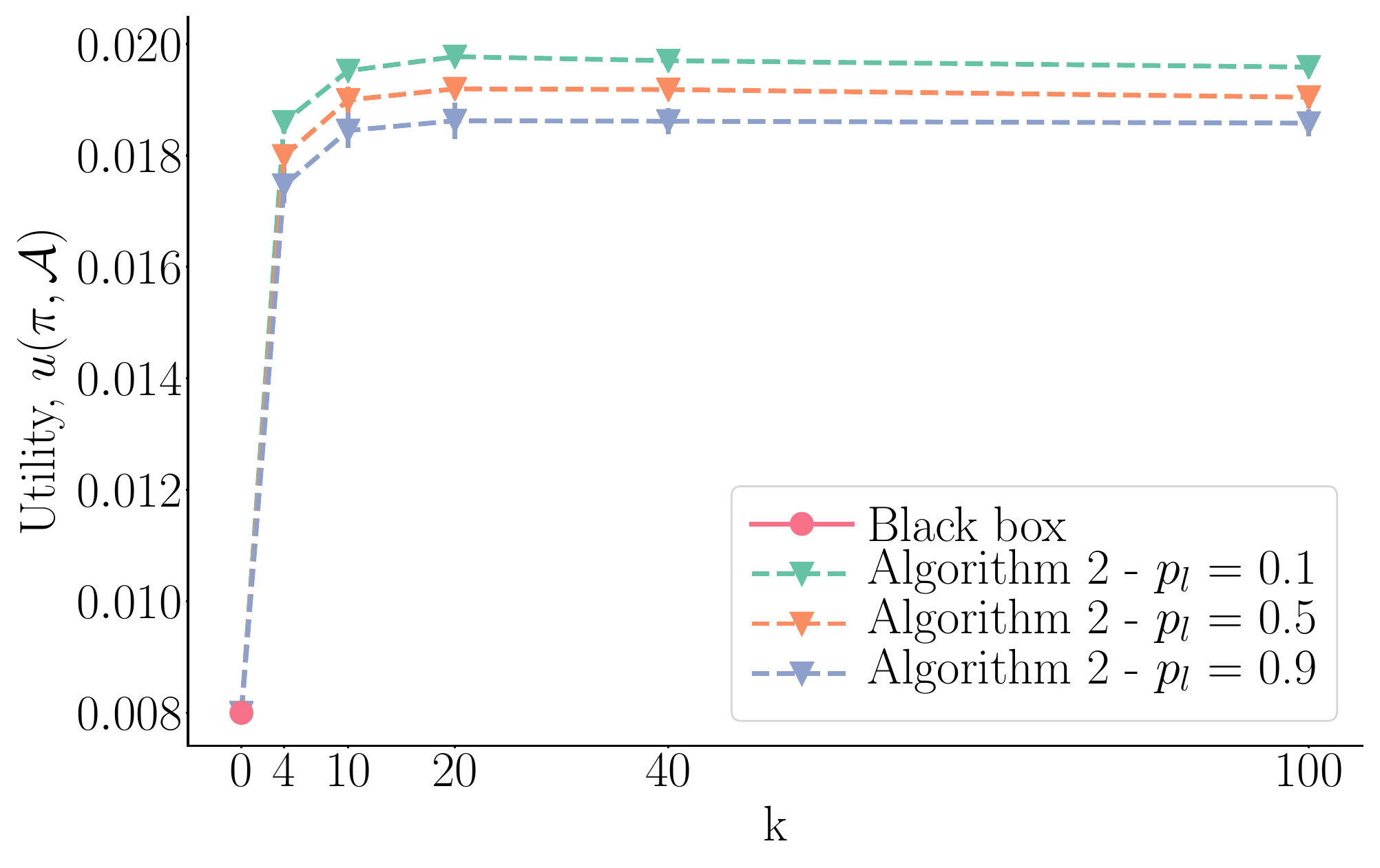}
    	}
     \caption{Number of counterfactual explanations and information leakage.
     Panel (a) shows the utility achieved by five types of decision policies and counterfactual explanations against the number of 
     counterfactual explanations $k$. 
     Panel (b) shows the utility achieved by Algorithm~\ref{alg:randomized} against the number of counterfactual explanations $k$
     for several values of the leakage probability $p_l$.
     In both panels, we use the lending dataset, the number of feature values is $m=400$, we set $\alpha = 2$, we repeat each 
     experiment involving randomization $20$ times.}
     \label{fig:real}
\end{figure}

\xhdr{Results} 
We start by comparing the utility achieved by each of the decision policies and counterfactual explanations in both datasets, for several values of the parameter 
$\alpha$, which is proportional to the difficulty of changing features.
%
%
%
%
Figure~\ref{fig:alphas} summarizes the results, which show that Algorithm~\ref{alg:greedy} and Algorithm~\ref{alg:randomized} consistently 
outperform all baselines and, as the cost of adapting to feature values with higher outcome values decreases (smaller $\alpha$), the 
competitive advantage by jointly optimizing the decision policy and the counterfactual explanations (Algorithm~\ref{alg:randomized}) 
grows significantly.
This competitive advantage is more apparent in the credit card dataset because it contains non actionable features (\eg, credit overdue 
counts) and, under the optimal decision policy in the non-strategic setting, it is difficult to incentivize individuals who receive a negative 
decision to improve by just optimizing the set of counterfactual explanations they receive.
For specific examples of counterfactual explanations provided by Algorithm~\ref{alg:greedy} and the minimum cost baseline, refer to Appendix~\ref{app:realistic}.

To understand the differences in utility caused by the two proposed algorithms, we measure the transportation of mass induced by the policies and counterfactual explanations used in Algorithm~\ref{alg:greedy} and~\ref{alg:randomized} in both datasets, as follows.
For each individual in the population whose best-response is to change her feature value, we record her outcome $P(y=1 \given \xb)$ before
and after the best response. 
Then, we discretize the outcome values using percentiles.
Figure~\ref{fig:transp} summarizes the results, which show several interesting insights.
In the lending dataset, we observe that a large portion of individuals do improve their outcome even if we only optimize the counterfactual
explanations (Panel (a)). 
In contrast, in the credit dataset, we observe that, if we only optimize the counterfactual explanations (Panel (c)), most individuals do not improve 
their outcome.
That being said, if we jointly optimize the decision policy and counterfactual explanations (Panels (b) and (d)), we are able to incentivize a large 
portion of individuals to self improve in both datasets.



\begin{figure}[t]
	\centering
	\subfloat[Rejected population by age]{
    		 \centering
    		 \includegraphics[scale=0.21]{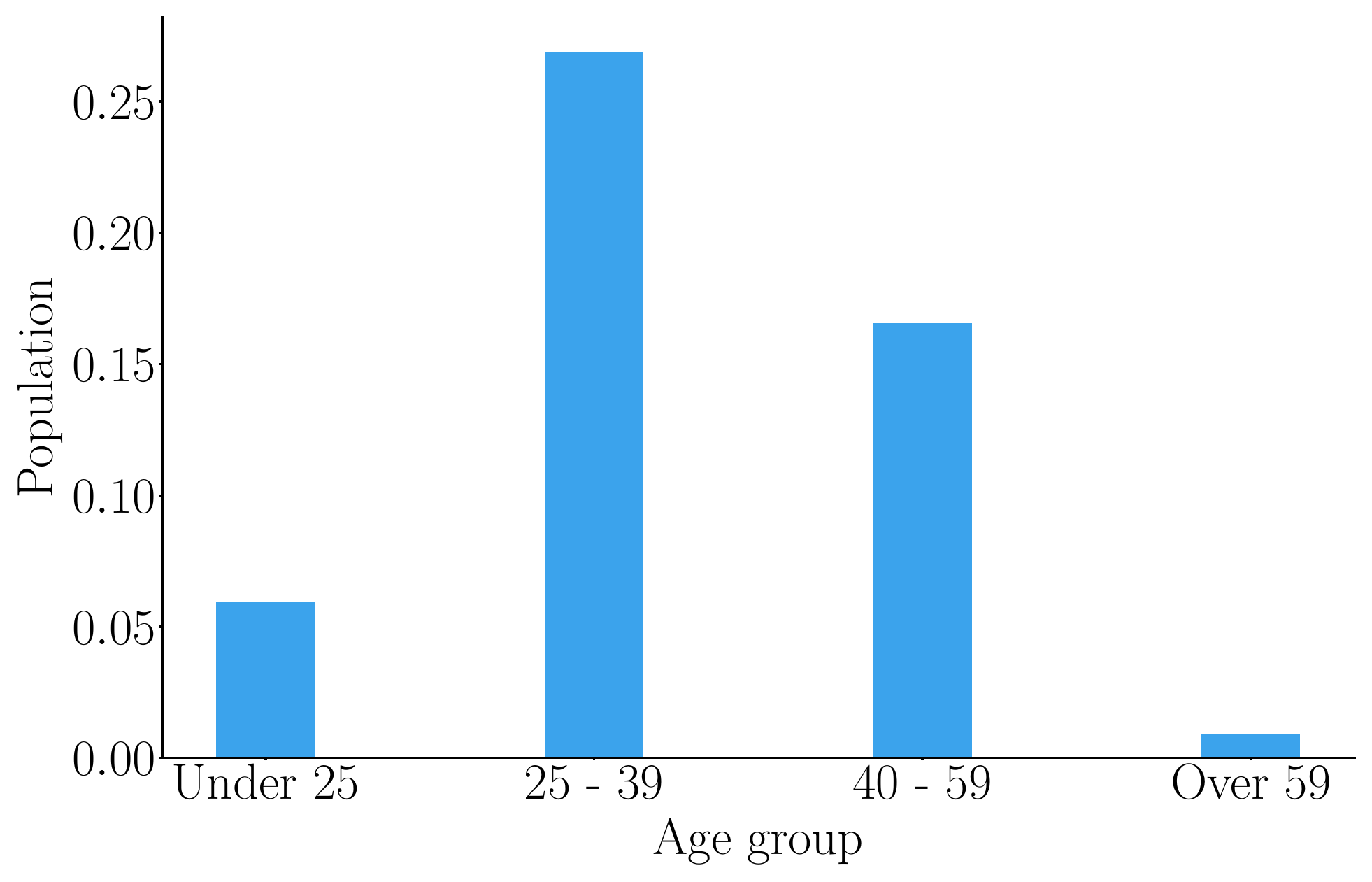}
    	}
    	\hspace{2mm}
    	\subfloat[Explanations by age]{
    		 \centering
    		 \includegraphics[scale=0.21]{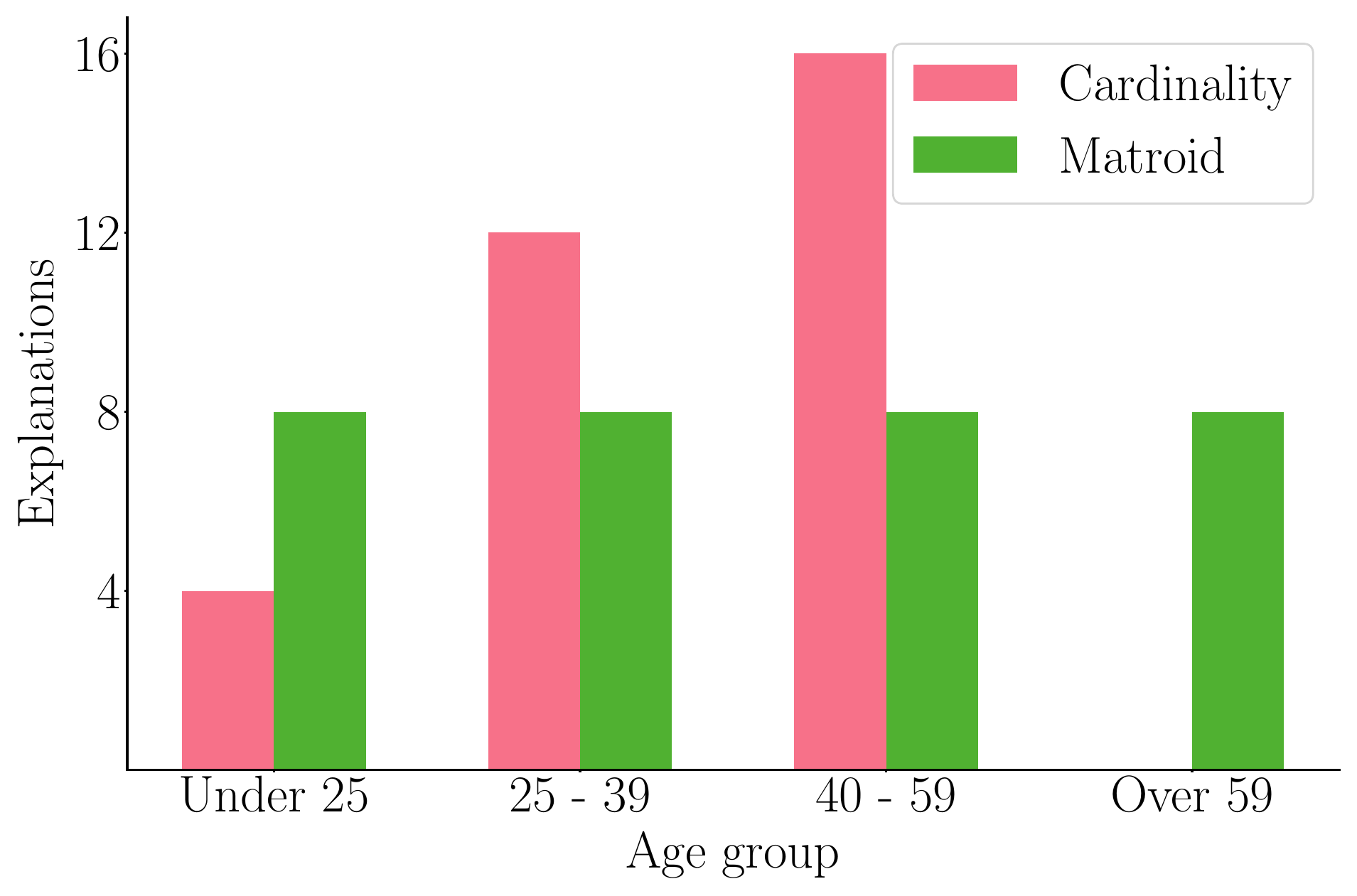}
    	}
    	\hspace{2mm}
    	\subfloat[Relative improvement by age]{
    		 \centering
    		 \includegraphics[scale=0.21]{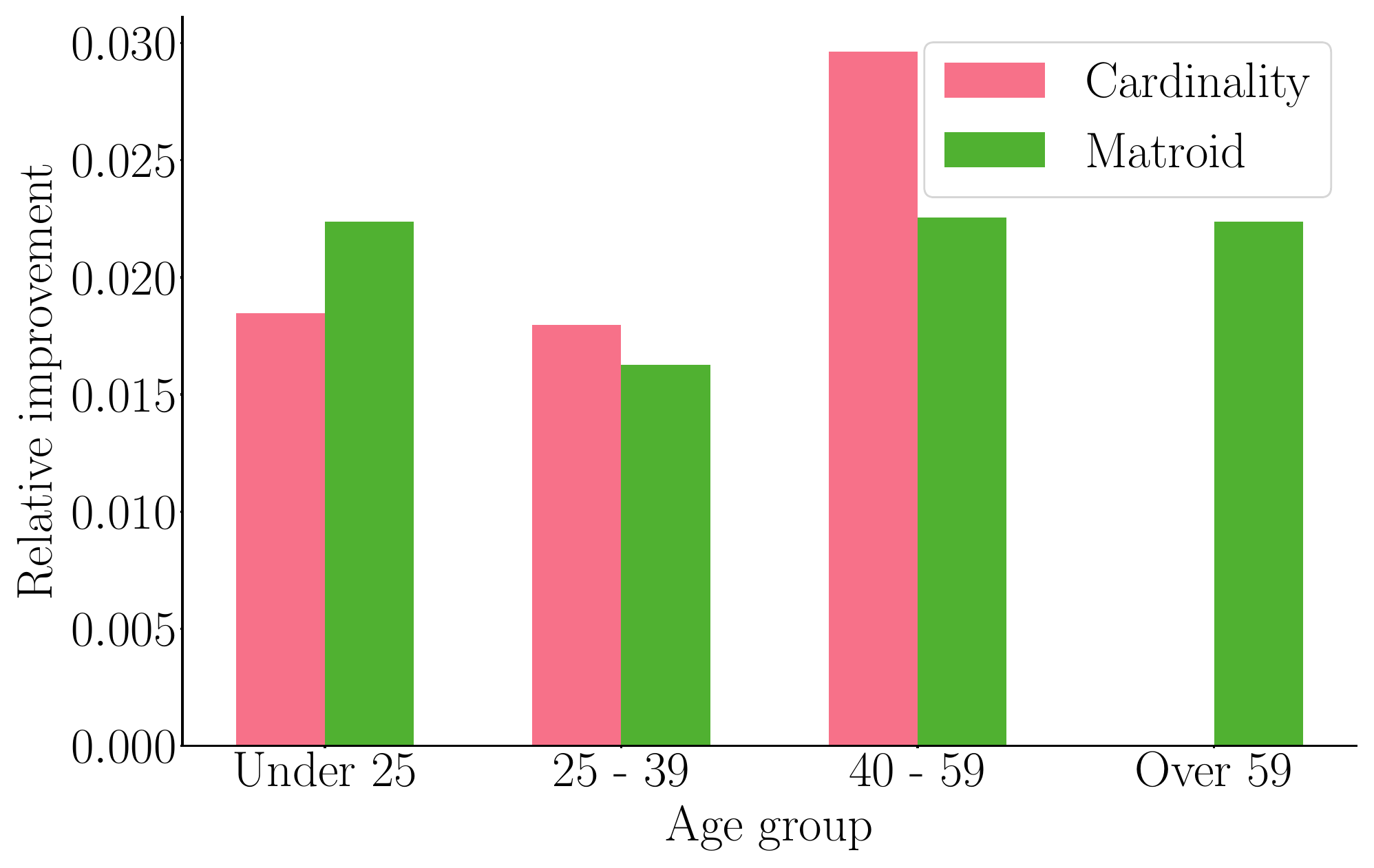}
    	}
     \caption{Increasing the diversity of the provided counterfactual explanations. 
     Panel (a) shows the population per age group, rejected by the optimal threshold policy in the non strategic setting.
     Panel (b) shows a comparison of the age distribution of counterfactual explanations in $\Acal$ produced by the greedy algorithm under 
     a cardinality and a matroid constraint. Panel (c) shows the relative improvement of each age group.
     In all panels, we use the credit dataset and we set $k = 32$ and $\alpha=2$.}
     \label{fig:fair}
\end{figure}

Next, we focus on the lending dataset and evaluate the sensitivity of our algorithms. First, we measure the influence that the number of 
counterfactual explanations has on the utility achieved by each of the decision policies and counterfactual explanations. 
As shown in Figure~\ref{fig:real}(a), our algorithms just need a small number of counterfactual explanations to provide significant gains in 
terms of utility with respect to all the baselines.
Second, we challenge the assumption that individuals do not share the counterfactual explanations they receive with other individuals with 
different feature values.
To this end, we assume that, given the set of counterfactual explanations $\Acal$ found by Algorithm~\ref{alg:randomized}, individuals with 
initial feature value $\xb$ receive the counterfactual explanation $\Ecal(\xb) \in \Acal$ given by Eq.~\ref{eq:best-counterfactual-explanations-rational} 
and, with probability $p_{l}$, they also receive an additional explanation $\Ecal'(\xb)$ picked at random from $\Acal$ and they follow the 
counterfactual explanation that benefits them the most. 
Figure~\ref{fig:real}(b) summarizes the results for several values of $p_l$ and number of counterfactual explanations, which show that the policies and 
explanations provided by Algorithm~\ref{alg:randomized} present a significant utility advantage even when the leakage probability $p_l$ is large.
Finally, we focus on the credit dataset and consider a scenario in which a bank aims not only to continue providing credit to the customers
that are more likely to repay but also provide explanations that incentivize individuals across all age groups to maintain their credit.
To this end, we incorporate a partition matroid constraint that ensures the counterfactual explanations are diverse across age groups, as 
described in Section~\ref{sec:matroids}, and use a slightly modified version of Algorithm~\ref{alg:greedy} to solve the constrained problem~\citep{nemhauser1978analysis}, which enjoys a $1/2$ approximation guarantee.
Figure~\ref{fig:fair} summarizes the results, which show that: 
(i) optimizing under a cardinality constraint leads to an unbalanced set of explanations, favoring the more populated age groups (25 to 59) while 
completely ignoring the recourse potential of individuals older than 60;
(ii) the relative group improvement, defined as $\sum_{\xb_i\in\Xcal_z\setminus\Pcal_\pi}P(\xb_i)[P(y\given\xb_j^i)-P(y\given\xb_i)] / \sum_{\xb_i\in\Xcal_z\setminus\Pcal_\pi}P(\xb_i)$, where $\Xcal_z$ is the set of feature values corresponding to age group $z$ and 
$\xb_j^i$ is the best response of individuals with initial feature value $\xb_i\in\Xcal_z$, is more balanced across age groups, showing 
that the matroid constraint can be used to generate counterfactual explanations that help the entire spectrum of the population 
to self-improve.
%

%% file: 070conclusions.tex
In this paper, we have designed several algorithms that allow us to find the decision policies and counterfactual explanations that maximize 
utility in a setting in which individuals who are subject to the decisions taken by the policies use the counterfactual explanations they receive to invest effort 
strategically.
Moreover, we have experimented with synthetic and real lending and credit card data and shown that the counterfactual explanations and decision
policies found by our algorithms achieve higher utility than several competitive baselines.

%
%
%

By uncovering a previously unexplored connection between strategic machine learning and interpretable machine learning, our work opens up
many interesting directions for future work.
For example, we have adopted a specific type of mechanism to provide counterfactual explanations (\ie, one feature value per individual using a Stackelberg formulation).
A natural next step would be to extend our analysis to other types of mechanisms fitting a variety of real-world applications.
Moreover, we have assumed that the cost individuals pay to change features is given. However, our algorithms would be more effective if we develop a methodology to reliably 
estimate the cost function from real observational (or interventional) data.
In our work, we have assumed that features take discrete values and individuals who are subject to the decisions do not share information between them. It would be interesting 
to lift these assumptions, extend our analysis to real-valued feature values, and develop decision policies and counterfactual explanations that are robust to information sharing 
between individuals (refer to Figure~\ref{fig:real}(c)).
Finally, by assuming that $P(y \given \xb)$ does not change after individuals best respond, we are implicitly assuming that  
there are not unobserved features that partially describe the true causal effect between the observed features $\xb$ and the 
outcome variable $y$. 
%
However, in practice, this assumption is likely to be violated and $P(y \given \xb)$ may change after individuals best respond, 
as recently noted by~\citet{miller2019strategic}. 
In this context, it would be very interesting to find counterfactual explanations that are robust to unmeasured confounding.
%

%% file: 080appendix.tex
\section{Further related work} \label{app:further-related-work}
\vspace{-3mm}
Our work builds upon previous work on interpretable machine learning, and strategic machine learning.

%
Most previous work on interpretable machine learning has focused on one of the two following types of explanations: feature-based 
explanations~\citep{ribeiro2016should, koh2017understanding, lundberg2017unified} or counterfactual explanations~\citep{wachter2017counterfactual, ustun2019actionable, karimi2019model, mothilal2019explaining}.
Feature-based explanations help individuals understand the importance each feature has on a particular prediction, typically through 
local approximation, while counterfactual explanations help them understand what features would have to change for a predictive 
model to make a positive prediction about them.
While there is not yet an agreement on what constitutes a \emph{good} post-hoc ex\-pla\-na\-tion in the literature on interpretable machine
learning, counterfactual explanations are gaining prominence because they place no constraints on the model complexity, do not require 
model disclosure, facilitate actionable recourse, and seem to automate compliance with the law~\citep{barocas2020hidden}.
Motivated by these desirable properties, our work focuses on counterfactual explanations and sheds light on the possibility of using explanations 
to increase the utility of a decision policy, uncovering a previously unexplored connection between interpretable machine learning and the nascent 
field of strategic machine learning.

Similarly as in our work, previous work on strategic machine learning also assumes that individuals may use knowledge, gained by transparency, to 
invest effort strategically in order to receive either a positive prediction~\citep{bruckner2011stackelberg, dalvi2004adversarial, dong2018strategic, hardt2016strategic, hu2019disparate, milli2019social, miller2019strategic, perdomo2020performative} or a beneficial decision~\citep{kleinberg2019classifiers, tabibian2020optimal}. 
However, none of this previous work focuses on finding (counterfactual) explanations and they assume full transparency---individuals who are subject 
to (semi)-automated decision making can observe the entire predictive model or the decision policy.
As a result, their formulation is fundamentally different and their technical contributions are orthogonal to ours.

\vspace{-3mm}
\section{Proofs}
\vspace{-3mm}
\subsection{Proof of Theorem~\ref{theor:np-hard}} \label{app:np-hard}
%
%
%
Consider an instance of the Set Cover problem with a set of elements $\Ucal=\{u_1,\ldots,u_n\}$ and a collection $\Scal=\{\Scal_1,\ldots, \Scal_m\}\subseteq 2^\Ucal$ such that $\bigcup_{i\in[m]}\Scal_i = \Ucal$.
In the decision version of the problem, given a constant $k$, we need to answer the question whether there are at most $k$ sets from the collection $\Scal$ such that their union is equal to $\Ucal$ or not.
With the following procedure, we show that any instance of that problem can be transformed to an instance of the problem of finding the optimal set of counterfactual
explanations, defined in Eq.~\ref{eq:optimal-counterfactual-explanations}, in polynomial time.

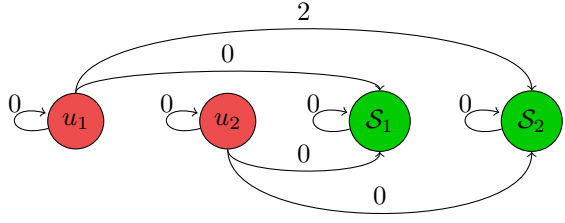
\begin{figure}
\centering
\begin{tikzpicture}
    \node[shape=circle,draw=black,fill=black!10!red!70] (A) at (0,0) {$u_1$};
    \node[shape=circle,draw=black,fill=black!10!red!70] (B) at (2,0) {$u_2$};
    \node[shape=circle,draw=black,fill=black!20!green] (C) at (4,0) {$\Scal_1$};
    \node[shape=circle,draw=black,fill=black!20!green] (D) at (6,0) {$\Scal_2$};

	\draw[->] (A) .. controls +(up:0.75cm) and +(up:0.75cm) .. node[above,sloped] {0} (C);
	\draw[->] (A) .. controls +(up:1.5cm) and +(up:1.5cm) .. node[above,sloped] {2} (D);
	\draw[->] (B) .. controls +(down:0.75cm) and +(down:0.75cm) .. node[above,sloped] {0} (C);
	\draw[->] (B) .. controls +(down:1.5cm) and +(down:1.5cm) .. node[above,sloped] {0} (D);
	
	\path (A) edge [loop left,above] node {0} (A);
	\path (B) edge [loop left,above] node {0} (B);
	\path (C) edge [loop left,above] node {0} (C);
	\path (D) edge [loop left,above] node {0} (D);
\end{tikzpicture}
\caption{Consider that $\Ucal=\{u_1, u_2\}$ and $\Scal=\{\Scal_1,\Scal_2\}$ with $\Scal_1=\{u_1,u_2\}$, $\Scal_2=\{u_2\}$. The red feature values have initial population $P(\xb)=1/2$, $\pi(\xb)=0$ and $P(y=1\given\xb)=\gamma$ while for the green feature values it is $P(\xb)=0$, $\pi(\xb)=1$ and $P(y=1\given\xb)=1$. The edges represent the cost between feature values corresponding to sets and their respective elements while all the non-visible pairwise costs are equal to 2.}
\label{fig:nphard}
\end{figure}

Consider $n+m$ feature values corresponding to the $n$ elements of $\Ucal$ and the $m$ sets of $\Scal$.
Moreover, denote the first $n$ feature values as $\xb_{u_1},\ldots,\xb_{u_n}$ and the remaining $m$ as $\xb_{\Scal_1},\ldots,\xb_{\Scal_m}$.
We set the decision maker'{}s parameter $\gamma$ to some positive constant less than $1$.
Then, we set the outcome probabilities $P(y=1|\xb_{u_i})=\gamma\ \forall i\in [n]$ and $P(y=1|\xb_{\Scal_i})=1\ \forall i\in [m]$ and the policy 
values $\pi(\xb_{u_i})=0\ \forall i\in [n]$ and $\pi(\xb_{\Scal_i})=1\ \forall i\in [m]$.
This way, the portion of utility the decision-maker obtains from the first $n$ feature values is zero, while the portion of utility she obtains from 
the remaining $m$ is proportional to $1-\gamma$.
Regarding the cost function, we set $c(\xb_{u_i},\xb_{\Scal_j})=0\ \forall (\xb_{u_i},\xb_{\Scal_j}) : u_i\in S_j$,
%
%
$c(\xb_{u_i},\xb_{u_i})=0\ \forall i\in[n]$, and all the remaining values of the cost function to $2$.
Finally, we set the initial feature value distribution to $P(\xb_{u_i})=\frac{1}{n}\ \forall i\in [n]$ and $P(\xb_{\Scal_i})=0\ \forall i\in [m]$.
A toy example of this transformation is presented in Figure~\ref{fig:nphard}.

In this setting, it easy to observe that an individual with initial feature value $\xb_{u_i}$ is always rejected at first and has the ability to move to a new feature value $\xb_{\Scal_j}$ recommended to her iff $c(\xb_{u_i},\xb_{\Scal_j})\leq 1 \Leftrightarrow u_i \in \Scal_j$. Also, we can easily see that the transformation of instances can be done in $\Ocal((m+n)^2)$ time.

Now, assume there exists an algorithm that optimally solves the problem of finding the optimal set of counterfactual explanations in polynomial time. Given the aforementioned 
instance and a maximum number of counterfactual explanations $k$, the utility $u(\pi,\Acal)$ achieved by the set of counterfactual explanations $\Acal$ the algorithm returns can fall into one of
the following two cases:

\begin{enumerate}
\item $u(\pi,\Acal)=1-\gamma$. This can happen only if all individuals, according to the induced distribution $P(\xb\given\pi,\Acal)$, have moved to some of the feature values $\xb_{\Scal_j}$, \ie, for all $\xb_{u_i}$ with $i \in [n]$, there exists $\xb_{\Scal_j}$ with $j \in [m]$ such that $\xb_{\Scal_j}\in\Acal \wedge c(\xb_{u_i},\xb_{\Scal_j})\leq 1$ with $|\Acal |\leq k$.
As a consequence, if we define $\Scal'=\{\Scal_j: \xb_{\Scal_j}\in\Acal\}$, it holds that for all $u_i$ with $i\in [n]$, there exists $\Scal_j$ with $j\in[m]$ such that $\Scal_j\in \Scal' \wedge u_i\in \Scal_j$ 
and therefore $\Scal'$ is a set cover with $|\Scal'|=|\Acal |\leq k$.

\item $u(\pi,\Acal)<1-\gamma$. This can happen only if every possible set of $k$ counterfactual explanations leaves the individuals of at least one feature value $\xb_{u_i}$ with a best-response of not following the counterfactual explanation they were given, \ie, for all $\Acal\subseteq \Pcal_\pi$ such that $|\Acal |\leq k$, there exists $\xb_{u_i}$ with $i\in[n]$ such that, for all $\xb_{\Scal_j}\in\Acal$, it holds that $c(\xb_{u_i},\xb_{\Scal_j})>1$.
Equivalently, it holds that for all $\Scal'\subseteq \Scal$ such that $|\Scal'|\leq k$, there exists $u_i$ with $i\in[n]$ such that for all $\Scal_j\in \Scal'$, it holds that $u_i\not\in \Scal_j$ and therefore there does not exist a set cover of size less or equal than $k$.
\end{enumerate}

The above directly implies that we can have a decision about any instance of the Set Cover problem in polynomial time, which is a contradiction unless $P=NP$.
This concludes the reduction and proves that the problem of finding the optimal set of counterfactual explanations for a given policy is NP-Hard.

\subsection{Proof of Proposition~\ref{prop:monsub}} \label{app:monsub}
It readily follows that the function $f$ is non-negative from the fact that, if the decision maker is rational, it holds that $\pi(\xb)=0$ for all $\xb \in \Xcal$ such 
that $P(y=1\given\xb)<\gamma$. 
%
%

Now, consider two sets $\Acal, \Bcal \subseteq \Pcal_\pi : \Acal \subseteq \Bcal$ and a feature value $\xb\in\Pcal_\pi\setminus \Bcal$.
Also, let $\Ecal_{\Scal}(\xb_i)$ be the counterfactual explanation given to the individuals with initial feature value $\xb_i$ under a set of counterfactual explanations $\Scal$.  
It is easy to see that the marginal difference $f(\Scal\cup\{\xb\})-f(\Scal)$ can only be affected by individuals with initial features $\xb_i$ such that $\xb_i\not\in\Pcal_\pi$, 
$\xb\in\Rcal(\xb_i)$ and $\xb=\Ecal_{S\cup\{\xb\}}(\xb_i)$.
Moreover, we can divide all of these individuals into two cases:
%
\begin{enumerate}
\item $\Rcal(\xb_i)\cap \Acal=\emptyset$: in this case, the addition of $\xb$ to $\Acal$ causes a change in their best-response from $\xb_i$ to $\xb$ contributing to the marginal difference of $f$ by a factor $P(\xb_i)[P(y=1\given\xb)-\gamma-\pi(\xb_i)(P(y=1\given\xb_i)-\gamma)]$. However, considering the marginal difference of $f$ under the set of counterfactual explanations $\Bcal$, three subcases are possible:
\begin{enumerate}
\item $\Ecal_{\Bcal}(\xb_i)\in\Rcal(\xb_i) \wedge P(y=1\given\Ecal_{\Bcal}(\xb_i))>P(y=1\given\xb)$: the contribution to the marginal difference of $f$ is zero.
\item $\Ecal_{\Bcal}(\xb_i)\in\Rcal(\xb_i) \wedge P(y=1\given\Ecal_{\Bcal}(\xb_i))\leq P(y=1\given\xb)$: the contribution to the marginal difference of $f$ is $P(\xb_i)[P(y=1\given\xb)-P(y=1\given\Ecal_{\Bcal}(\xb_i))]$. Since $\pi$ is outcome monotonic, $\Ecal_{\Bcal}(\xb_i)\in\Pcal_\pi$ and $\xb_i\not\in\Pcal_\pi$, it holds that 
\begin{flalign*}
\quad \,\,  &P(y=1\given\Ecal_{\Bcal}(\xb_i))\geq P(y=1\given\xb_i)\Rightarrow &&\\
&P(y=1\given\Ecal_{\Bcal}(\xb_i))-\gamma\geq P(y=1\given\xb_i)-\gamma>\pi(\xb_i)[P(y=1\given\xb_i)-\gamma].
\end{flalign*}
Therefore, it readily follows that 
\begin{flalign*}
\quad \,\, &P(\xb_i)[P(y=1\given\xb)-P(y=1\given\Ecal_{\Bcal}(\xb_i))]< &&\\
& \qquad P(\xb_i)[P(y=1\given\xb)-\gamma-\pi(\xb_i)(P(y=1\given\xb_i)-\gamma)].
\end{flalign*}
\item $\Rcal(\xb_i)\cap \Bcal=\emptyset$: the contribution to the marginal difference of $f$ is $P(\xb_i)[P(y=1\given\xb)-\gamma-\pi(\xb_i)(P(y=1\given\xb_i)-\gamma)]$.
\end{enumerate}
\item $\Rcal(\xb_i)\cap \Acal\neq\emptyset \wedge P(y=1\given\xb)>P(y=1\given\Ecal_{\Acal}(\xb_i))$: In this case, the addition of $\xb$ to $\Acal$ causes a change in their best-response from $\Ecal_{\Acal}(\xb_i)$ to $\xb$ contributing to the marginal difference of $f$ by a factor $P(\xb_i)[P(y=1\given\xb)-P(y=1\given\Ecal_{\Acal}(\xb_i))]$. Considering the marginal difference of $f$ under the set of counterfactual explanations $\Bcal$, two subcases are possible:
\begin{enumerate}
\item $\Ecal_{\Bcal}(\xb_i)\in\Rcal(\xb_i) \wedge P(y=1\given\Ecal_{\Bcal}(\xb_i))>P(y=1\given\xb)$: the contribution to the marginal difference of $f$ is zero.
\item $\Ecal_{\Bcal}(\xb_i)\in\Rcal(\xb_i) \wedge P(y=1\given\Ecal_{\Bcal}(\xb_i))\leq P(y=1\given\xb)$. Then, the contribution of those individuals to the marginal difference of $f$ 
is $P(\xb_i)[P(y=1\given\xb)-P(y=1\given\Ecal_{\Bcal}(\xb_i))]$. Since $\Acal\subseteq \Bcal$ and $\Rcal(\xb_i)\cap \Acal\neq\emptyset$, it readily follows that
\begin{flalign*}
\quad \,\, &P(y=1\given\Ecal_{\Bcal}(\xb_i))\geq P(y=1\given\Ecal_{\Acal}(\xb_i)) \Rightarrow &&\\
&P(\xb_i)[P(y=1\given\xb)-P(y=1\given\Ecal_{\Acal}(\xb_i))]\geq &&\\
& \qquad P(\xb_i)[P(y=1\given\xb)-P(y=1\given\Ecal_{\Bcal}(\xb_i))].
\end{flalign*}
\end{enumerate}
\end{enumerate}

Finally, because $\Acal \subseteq \Bcal$, we can conclude that $f(\Bcal \cup\{\xb\})-f(\Bcal)\neq 0 \Rightarrow f(\Acal\cup\{\xb\})-f(\Acal)\neq 0$ and therefore the aforementioned 
cases are sufficient.
Combining all cases, we can see that the contribution of each individual to the marginal difference of $f$ is always greater or equal under the set of counterfactual explanations 
$\Acal$ than under the set of counterfactual explanations $\Bcal$. As a direct consequence, it follows that $f$ is submodular.
Additionally, we can easily see that this contribution is always greater or equal than zero, leading to the conclusion that $f$ is also monotone.

\subsection{Proof of Proposition~\ref{prop:uniq}} \label{app:uniq}
By definition, since $\Acal\subseteq\Pcal_{\pi_\Acal^*}$, it readily follows that $\pi_\Acal^*(\xb)=1$ for all $\xb \in \Acal$.
To find the remaining values of the decision policy, we first observe that, for each $\xb \notin \Acal$, the value of the decision policy $\pi_\Acal^*(\xb)$ does not affect 
the best-responses of the individuals with initial feature values $\xb' \neq \xb$. 
%
%
As a result, we can just set $\pi_\Acal^*(\xb)$ for all $\xb \notin \Acal$ independently for each feature value $\xb$ such that the best-response of the respective individuals 
is the one that contributes maximally to the overall utility.

First, it is easy to see that, for all $\xb \notin \Acal$ such that $P(y=1\given\xb)<\gamma$, we should set $\pi_\Acal^*(\xb)=0$. 
%
%
%
%
%
Next, consider the feature values $\xb \notin \Acal$ such that $P(y=1\given\xb)\geq\gamma$. Here, we distinguish two cases.
If there exists $\xb' \in \Acal$ such that $c(\xb,\xb') \leq 1 \wedge P(y=1\given\xb')>P(y=1\given\xb)$, then, if the individuals move
to that $\xb'$, the corresponding contribution to the utility will be higher. 
Moreover, the value of the decision policy that maximizes their region of adaption (and thus increases their chances of moving to 
$\xb'$) is clearly $\pi_{\Acal}^*(\xb) = 0$.
%
%
%
%
If there does not exist $\xb' \in \Acal$ such that $c(\xb,\xb') \leq 1 \wedge P(y=1\given\xb')>P(y=1\given\xb)$, then, the contribution of
the corresponding individuals to the utility will be higher if they keep their initial feature values. Moreover, the value of the decision
policy that will maximize this contribution will be clearly $\pi_\Acal^*(\xb)=1$.
%
%
%
%
%
%
%

\subsection{Proof of Proposition~\ref{prop:nonmon}} \label{app:nonmon}
It readily follows that the function $h$ is non-negative from the fact that, if the decision maker is rational, $\pi(\xb)=0$ for all $\xb \in \Xcal$ such that 
$P(y = 1\given\xb)<\gamma$.

%

Next, consider two sets $\Acal, \Bcal \subseteq \Ycal$ such that $\Acal \subseteq \Bcal$ and a feature value $\xb\in\Ycal\setminus \Bcal$.
Also, let $\Ecal_\Scal(\xb_i)$ be the counterfactual explanation given to the individuals with initial feature value $\xb_i$ under a set of counterfactual 
explanations $\Scal$.  
%
%
%
Then, it is clear that the marginal difference $h(\Scal\cup\{\xb\})-h(\Scal)$ only depends on individuals with initial features $\xb_i$ such that either $1-c(\xb_i,\xb)\geq 0$ 
and $\xb=\Ecal_{\Scal\cup\{\xb\}}(\xb_i)$ or $\xb_i=\xb$.
Moreover, if $1-c(\xb_i,\xb)\geq 0$ and $\xb=\Ecal_{\Scal\cup\{\xb\}}(\xb_i)$, the contribution to the marginal difference is positive and, if $\xb_i=\xb$, the contribution to
the marginal difference is negative.
%
%
%

%
Consider first the individuals with initial features $\xb_i$ such that $1-c(\xb_i,\xb)\geq 0$ and $\xb=\Ecal_{\Acal\cup\{\xb\}}(\xb_i)$. We can divide all of these individuals into 
three cases:
\begin{enumerate}
\item $\pi_\Bcal(\xb_i)=0$: in this case, $\xb_i\not\in \Bcal$ and the individuals change their best-response from $\Ecal_\Bcal(\xb_i)$ to $\xb$. Moreover, under the set of counterfactual 
explanations $\Acal$, their best-response is either $\xb_i$ or $\Ecal_\Acal(\xb_i)$ and it changes to $\xb$. 
Then, using a similar argument as in the proof of proposition~\ref{prop:monsub}, we can conclude that the contribution of the individuals to the marginal difference is greater or equal under the 
set of counterfactual explanations $\Acal$ than under $\Bcal$.
%

\item $\pi_\Bcal(\xb_i)=1 \wedge \pi_\Acal(\xb_i)=0$: in this case, $\xb_i\not\in \Acal$ and $\xb_i\in \Bcal$. Therefore, under the set of counterfactual explanations $\Acal$, the individuals'{} 
best-response changes from $\Ecal_\Acal(\xb_i)$ to $\xb$ and there is a positive contribution to the marginal difference while, under $\Bcal$, the individuals{}' best response does not change 
and the contribution to the marginal difference is zero.

\item $\pi_\Bcal(\xb_i)=1 \wedge \pi_\Acal(\xb_i)= 1$: in this case, $\xb_i \not \in \Bcal$. Therefore, the best-response changes from $\xb_i$ to $\xb$ under both sets of counterfactual explanations
and there is an equal positive contribution to the marginal difference.
\end{enumerate}
Now, consider the individuals with initial features $\xb_i$ such that $\xb_i = \xb$. We can divide all of these individuals also into three cases:
\begin{enumerate}
\item $\pi_\Acal(\xb)=\pi_\Bcal(\xb)=0$: in this case, under both sets of counterfactual explanations, the counterfactual explanation $\xb$ changes the value of the decision policy to
$\pi_{\Acal \cup \{\xb\}}(\xb)=\pi_{\Bcal \cup \{\xb\}}(\xb)=1$. Moreover, the contribution to the marginal difference is less negative under the set of counterfactual explanations 
$\Acal$ than under $\Bcal$ since $P(y=1\given\Ecal_\Acal(\xb)) \leq P(y=1\given\Ecal_\Bcal(\xb))$ and thus $P(\xb)[P(y = 1\given\xb) - P(y = 1\given\Ecal_\Acal(\xb))] \geq P(\xb)[P(y=1\given\xb) - P(y=1\given\Ecal_\Bcal(\xb))]$.
\item $\pi_\Acal(\xb)=1 \wedge \pi_\Bcal(\xb)=0$: in this case, under the set of counterfactual explanations $\Acal$, the individuals'{} best response does not change and thus the
contribution to the marginal difference is zero and, under the set of counterfactual explanations $\Bcal$, their best-response changes from $\Ecal_\Bcal(\xb)$ to $\xb$ and thus there 
is a negative contribution to the marginal difference \ie, $P(\xb)[P(y=1\given\xb) - P(y=1\given\Ecal_\Bcal(\xb))] < 0$.
\item $\pi_\Acal(\xb)=\pi_\Bcal(\xb)=1$: in this case, under both sets of counterfactual explanations, the individuals'{} best response does not change and thus the contribution to the
marginal difference is zero.
\end{enumerate}

As a direct consequence of the above observations, it readily follows that $h(\Acal\cup\{\xb\})-h(\Acal)\geq h(\Bcal\cup\{\xb\})-h(\Bcal)$ and therefore the function $h$ is submodular.

However, in contrast with Section~\ref{sec:fixed}, the function $h$ is non-monotone since it can happen that the negative marginal contribution exceeds the positive one. For example, consider the following 
instance of the problem, where $\xb\in\{1,2,3\}$ with $\gamma=0.1$:
\begin{align*}
P(\xb) &= 0.1\, \II(\xb=1) + 0.8\, \II(\xb=2) + 0.1\, \II(\xb=3), \\
P(y = 1 \given \xb) &= 1.0\, \II(\xb=1) + 0.5\, \II(\xb=2) + 0.4\, \II(\xb=3),
\end{align*}
and
\begin{align*}
  c(\xb_i, \xb_j) &= \begin{bmatrix} 0.0 & 0.2 & 0.3 \\
                                  0.3 & 0.0 & 0.7 \\
                                  0.4 & 0.5 & 0.0
                      \end{bmatrix}.
\end{align*}

Assume there is a set of counterfactual explanations $\Acal=\{1\}$.
Then, the optimal policy is given by $\pi_\Acal^*(1)=1, \pi_\Acal^*(2)=0, \pi_\Acal^*(3)=0$ inducing a movement from feature values $2,3$ to feature value $1$, giving a utility equal 
to $0.9$.
Now, add $\xb=2$ to the set of counterfactual explanations \ie, $\Acal=\{1,2\}$.
Then, the optimal policy is given by $\pi_\Acal^*(1)=1, \pi_\Acal^*(2)=1, \pi_\Acal^*(3)=0$ inducing a movement from feature value $3$ to feature value $1$, giving a lower utility, 
equal to $0.5$.
Therefore, the function $h$ is non-monotone.
\begin{algorithm}[t]
\renewcommand{\algorithmicrequire}{\textbf{Input:}}
\renewcommand{\algorithmicensure}{\textbf{Output:}}
\caption{Standard greedy algorithm~\citep{nemhauser1978analysis}}
\label{alg:greedy}
\begin{algorithmic}[1]
\small
\REQUIRE Ground set of counterfactual explanations $\Pcal_{\pi}$, parameter $k$ and utility function $f$
\ENSURE Set of counterfactual explanations $\Acal$
\STATE $\Acal \leftarrow \varnothing$
\WHILE{$|\Acal| \leq k$}
\STATE $\xb^* \leftarrow \text{argmax}_{\xb \in \Pcal_{\pi} \backslash \Acal} f(\Acal\cup\{\xb\})-f(\Acal)$
\STATE $\Acal \leftarrow \Acal \cup \{ \xb^* \}$
\ENDWHILE \\
\STATE \mbox{\bf return} $\Acal$
\end{algorithmic}
\end{algorithm}
\begin{algorithm}[t]
\renewcommand{\algorithmicrequire}{\textbf{Input:}}
\renewcommand{\algorithmicensure}{\textbf{Output:}}
\caption{Randomized algorithm by~\citet{buchbinder2014submodular}}
\label{alg:randomized}
\begin{algorithmic}[1]
\small
\REQUIRE Ground set of counterfactual explanations $\Ycal$, parameter $k$ and utility function $f$
\ENSURE Set of counterfactual explanations $\Acal$
\STATE $\Acal \leftarrow \varnothing$
\WHILE{$|\Acal| \leq k$}
	\STATE $\Bcal \leftarrow$ GetTopK$(\Ycal, \Acal, f)$
	\STATE $\xb^* \sim \Bcal$
	\STATE $\Acal \leftarrow \Acal \cup \{ \xb^* \}$
\ENDWHILE \\
\STATE \mbox{\bf return} $\Acal$
\end{algorithmic}
\end{algorithm}

\vspace{-3mm}
\section{Additional details on the standard greedy algorithm and the randomized algorithm by~\citet{buchbinder2014submodular}} 
\label{app:algorithms}
\vspace{-3mm}
To enjoy a $1/e$ approximation guarantee, Algorithm~\ref{alg:randomized} requires that there are $2k<m$ candidate feature values whose marginal 
contribution to any set is zero.
In our problem, this can be trivially satisfied by adding $2k$ feature values $\xb$ to $\Xcal$ such that $P(y=1\given\xb)=\gamma$, $P(\xb)=0$ and $c(\xb,\xb_j)=c(\xb_j,\xb)=2\ \forall\xb_j\in\Xcal$.
If the algorithm adds some of those counterfactual explanations to the set $\Acal$, it is easy to see that we can ignore them without causing any difference in utility or best-responses. 

\vspace{-3mm}
\section{Jointly optimizing the decision policy and the counterfactual explanations} \label{app:pushed}
\begin{figure}[h]
\captionsetup[subfigure]{labelformat=empty}
	\centering
	\subfloat[Non-strategic policy]{
    		 \centering
    		 \includegraphics[scale=0.27]{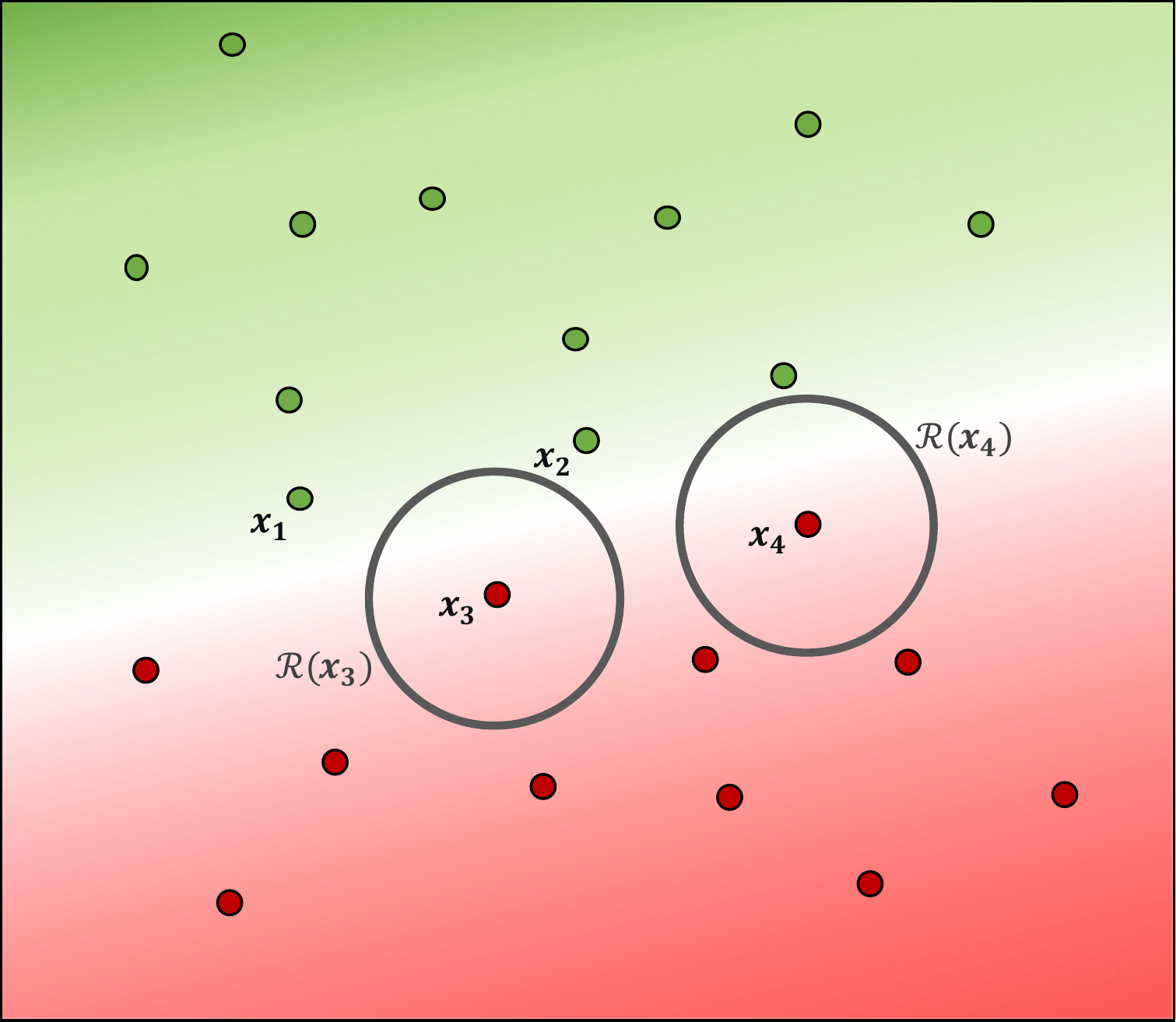}
    	}
    	\hspace{1mm}
	    	\subfloat[]{\centering
    		 \includegraphics[scale=0.27]{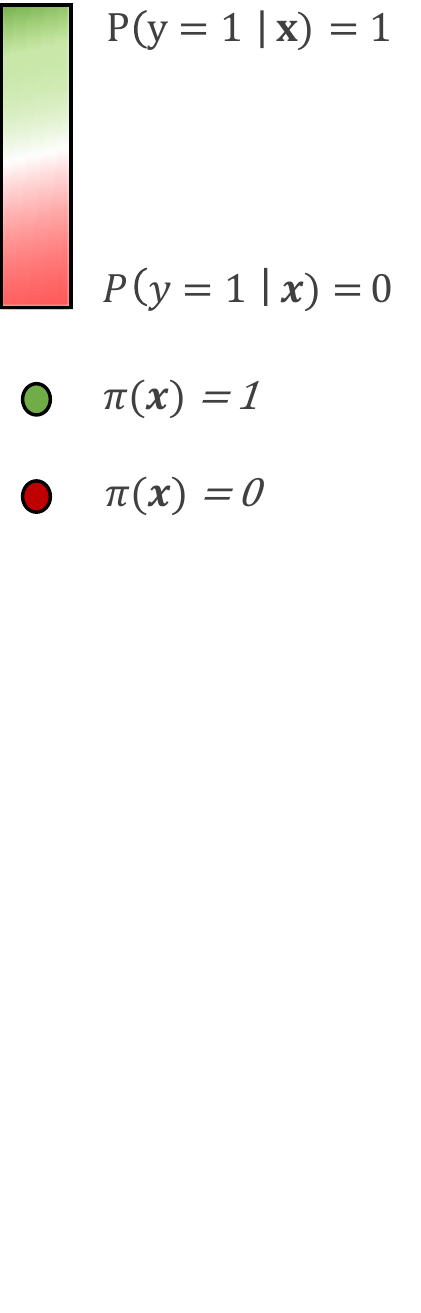}}
    	\hspace{1mm}
    	\subfloat[Strategic policy]{
    		 \centering
    		 \includegraphics[scale=0.27]{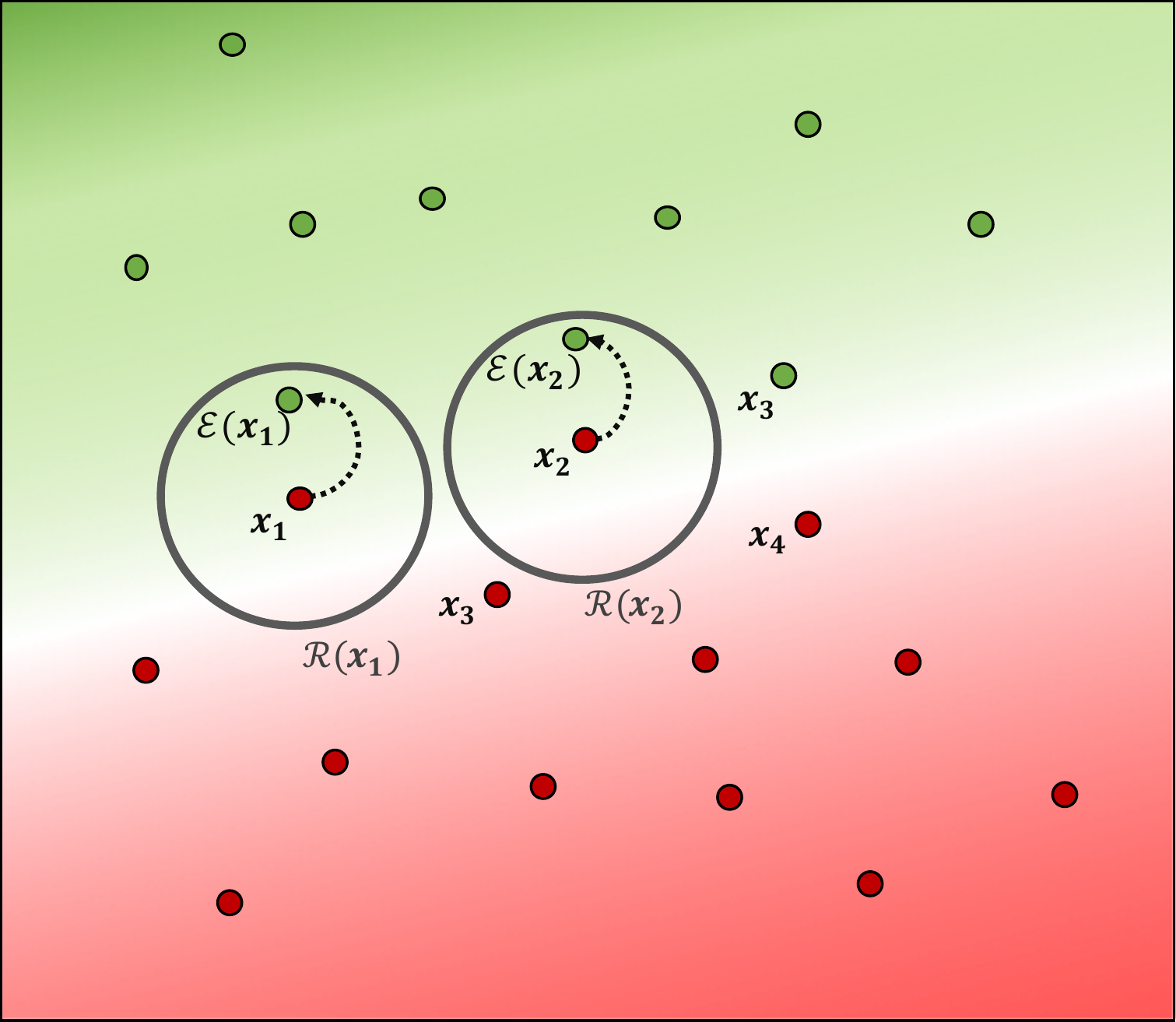}
	}
     \caption{Jointly optimizing the decision policy and the counterfactual explanations can offer additional gains. 
     The left panel shows the optimal (deterministic) decision policy $\pi$ under non-strategic behavior, as given by Eq.~\ref{eq:dtr}. Here, there does not exist a set of 
     counterfactual explanations $\Acal \in \Pcal_{\pi}$ that increases the utility of the policy. This happens because the area of adaption of $\xb_3$ and $\xb_4$ 
     does not include any feature value that receives a positive decision.
     The right panel shows the decision policy and counterfactual explanations that are (jointly) optimal in terms of utility, as given by Eq.~\ref{eq:optimal-policy-and-counterfactual-explanations}. Here, the individuals with feature values $\xb_1$ and $\xb_2$
     receive $\Ecal(\xb_1)$ and $\Ecal(\xb_2)$, respectively, as counterfactual explanations. Since these explanations are within their areas of adaptation $\Rcal(\xb_1)$ and $\Rcal(\xb_2)$, 
     they change their initial feature values in order to receive a positive decision.}
     %
     \label{fig:pushed}
\end{figure}

\vspace{-3mm}
Figure~\ref{fig:pushed} shows that, by jointly optimizing both the decision policy and the counterfactual explanations, we may obtain an additional gain in terms of utility in 
comparison with just optimizing for the set of counterfactual explanations given the optimal decision policy in a non-strategic setting.

\vspace{-3mm}
\section{Experiments on Synthetic Data} \label{app:synthetic}
\vspace{-3mm}
\xhdr{Experimental setup} For simplicity, we consider feature values $\xb\in\{0,\ldots,m-1\}$ and $P(\xb=i)=p_i/\sum_j p_j$ where $p_i$ is sampled from a Gaussian distribution $N(\mu=0.5,\sigma=0.1)$ truncated from below at zero.
We also sample $P(y=1\given\xb)\sim U[0,1]$, $c(\xb_i,\xb_j)\sim U[0,1]$ for $50\%$ of all pairs and $c(\xb_i,\xb_j)=2$ for the rest.
Finally, we set $\gamma=0.3$.
In this section, we compare the utility achieved by our explanation methods with the same baselines we used on real data.

\begin{figure}[t]
	\centering
	\subfloat[Utility vs. \# feature values]{
    		 \centering
    		 \includegraphics[scale=0.21]{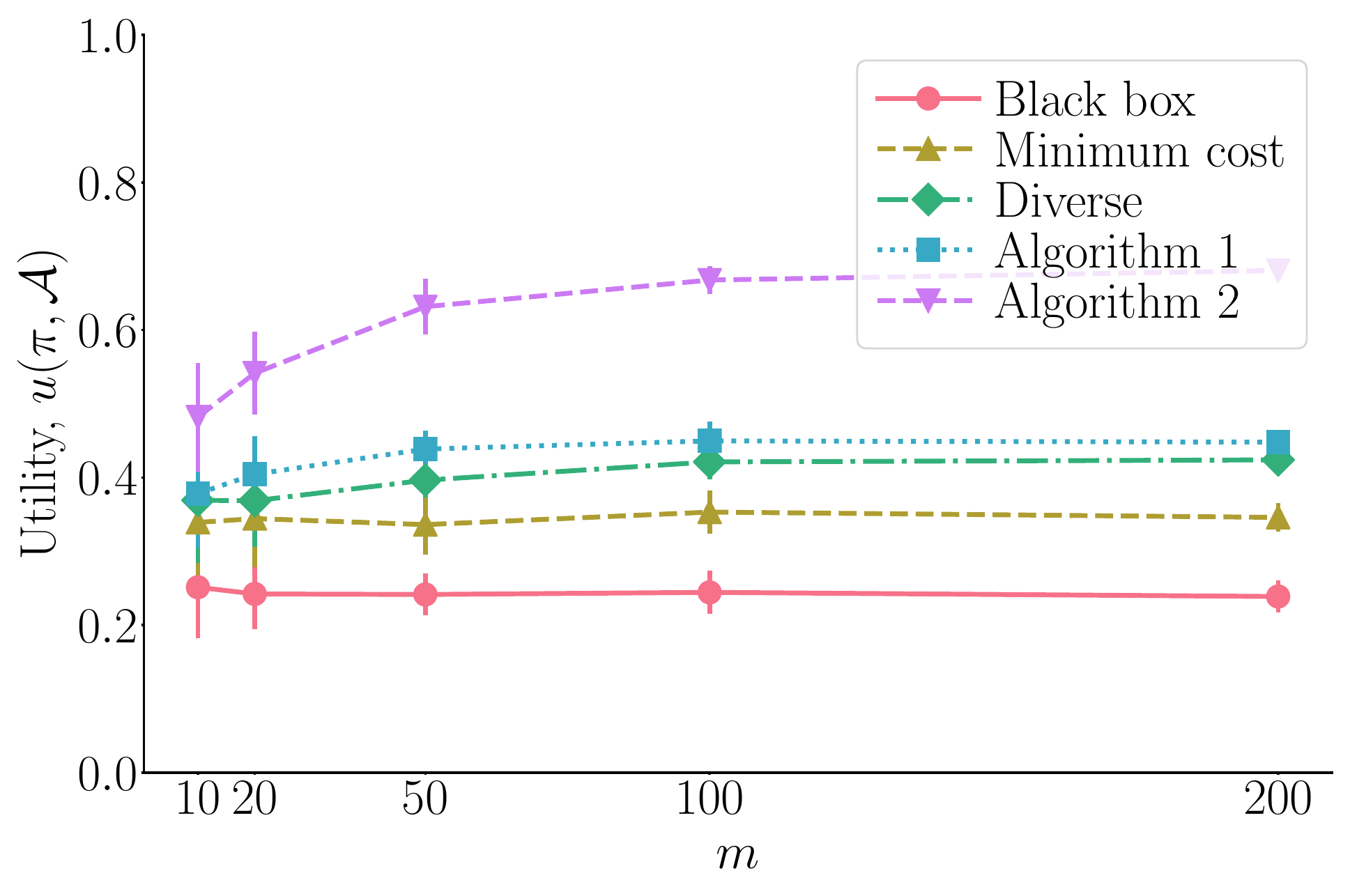}
    	}
    	\hspace{2mm}
    	\subfloat[Utility vs. \# explanations]{
    		 \centering
    		 \includegraphics[scale=0.21]{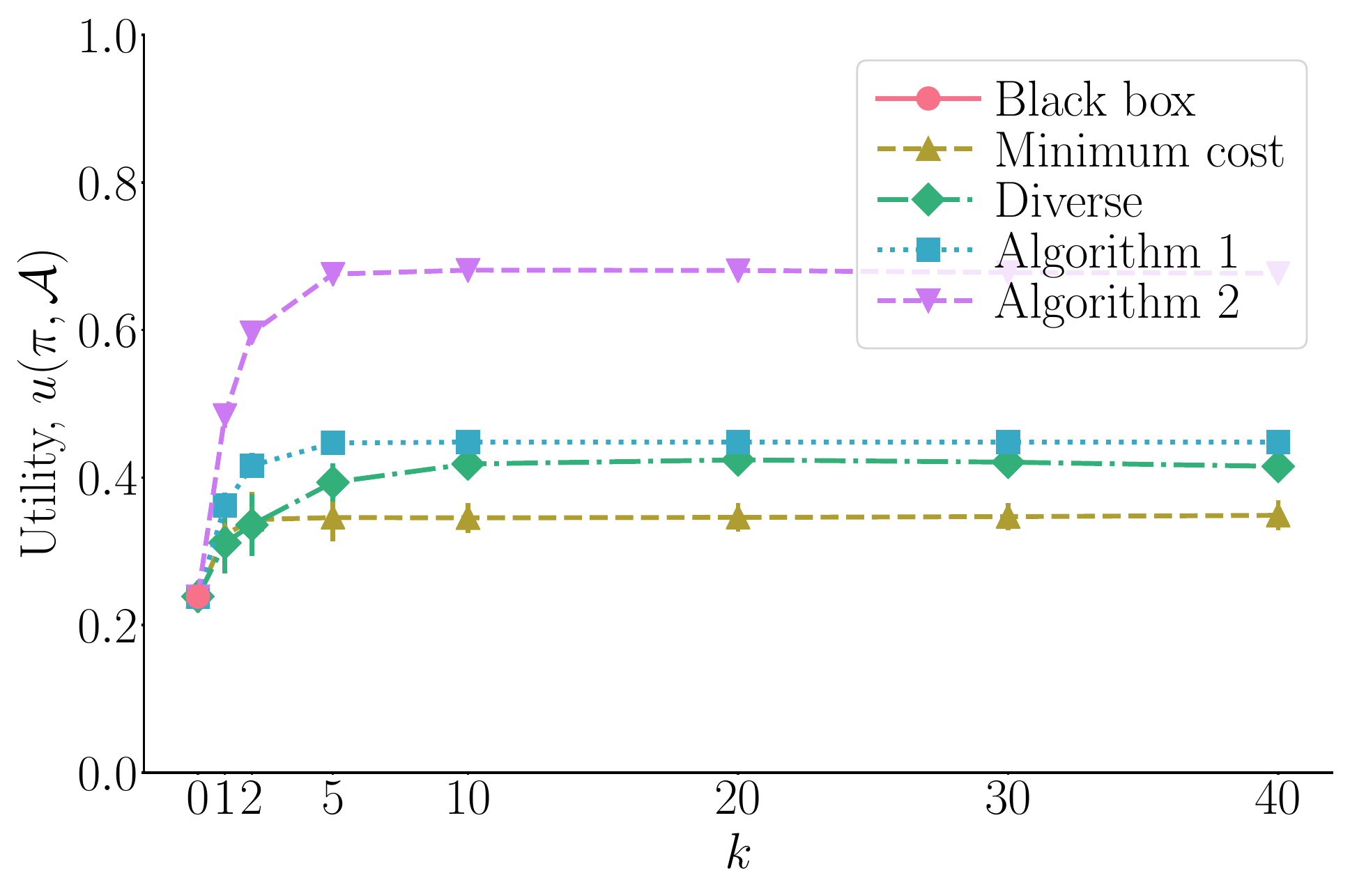}
    	}
    	\hspace{2mm}
    	\subfloat[Individual cost vs. \# explanations]{
    		 \centering
    		 \includegraphics[scale=0.21]{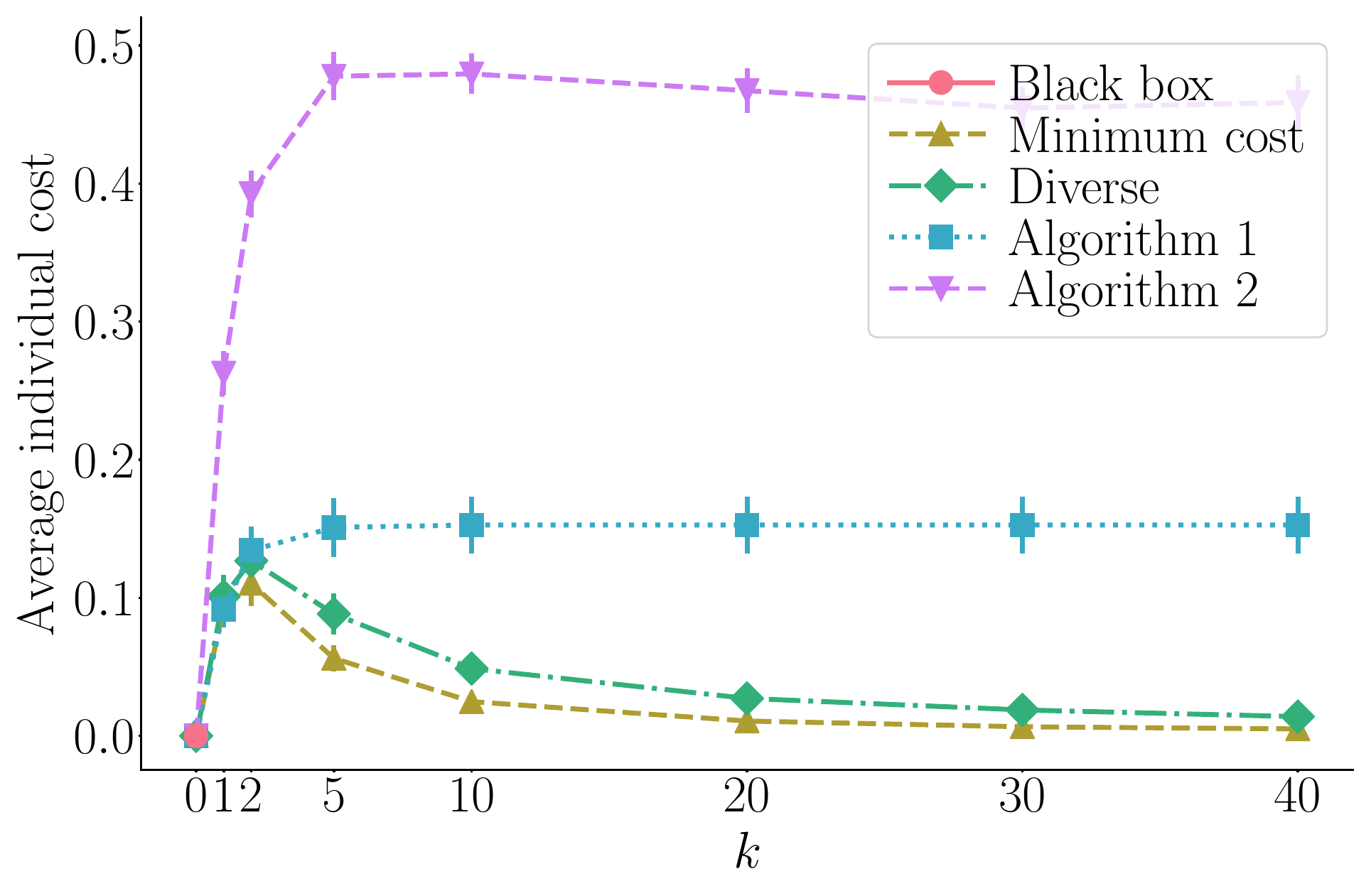}
    	}
     \caption{Results on synthetic data. Panels (a) and (b) show the utility achieved by six types of decision policies and counterfactual explanations against the total number 
     of feature values $m$ and the number of counterfactual explanations $k$, respectively. 
     Panel (c) shows the average cost individuals had to pay to change from their initial features to the feature value of the counterfactual explanation they receive under the
     same five types of decision policies and counterfactual explanations. 
     In Panel (a), we set $k = 0.1 m$ and, in Panels (b) and (c), we set $m = 200$. 
     In all panels, we repeat each experiment $20$ times.}
     \label{fig:synth}
\end{figure}

\xhdr{Results} Figures~\ref{fig:synth}(a,b) show the utility achieved by each of the decision policies and counterfactual explanations for several numbers of feature values $m$ and counterfactual explanations $k$. 
We find several interesting insights: 
(i) the decision policies given by Eq.~\ref{eq:expol} and the counterfactual explanations found by Algorithm~\ref{alg:randomized} beat all other alternatives by large margins 
across the whole spectrum, showing that jointly optimizing the decision policy and the counterfactual explanations offer clear additional gains; 
(ii) the counterfactual explanations found by Algorithms~\ref{alg:greedy} and~\ref{alg:randomized} provide higher utility gains as the number of feature values increases and 
thus the search space of counterfactual explanations becomes larger; and,
(iii) a small number of counterfactual explanations is enough to provide significant gains in terms of utility with respect to the optimal decision policy without counterfactual
explanations.

%
%
Figure~\ref{fig:synth}(c) shows the average cost individuals had to pay to change from their initial features to the feature value of the counterfactual explanation they 
receive. 
As one may have expected, the results show that, under the counterfactual explanations of minimum cost (Minimum cost and Diverse), the individuals invest less effort
to change their initial features and the effort drops as the number of counterfactual explanations increases. 
In contrast, our methods incentivize the individuals to achieve the highest self-improvement, particularly when we jointly optimize the decision policy and the counterfactual
explanations. 
%
%


\vspace{-3mm}
\section{Additional details on the experiments on real data} \label{app:real_data}
\vspace{-2mm}

\subsection{Feature representation \& preprocessing steps} \label{app:feature-representation}

For each applicant in the lending dataset, the label $y$ indicates whether an applicant fully pays a loan ($y=1$) or ends up to a default/charge-off ($y=0$) 
and the features $\xb$ are:
 \begin{itemize}
    \item Loan Amount: The amount that the applicant initially requested.
    \item Employment Length: How long the applicant has been employed.
    \item Debt to Income Ratio: The ratio between the applicant's financial debts and her average income.
    \item FICO Score: The applicant's FICO score, which is a credit score based on consumer credit files. The FICO scores are in the range of 300-850 and 
 the average of the high and low range for the FICO score of each applicant has been used for this study.
 	\item Annual Income: The declared annual income of the applicant.
 \end{itemize}
Here, we assume that all of the aforementioned features are \textit{actionable}, meaning that an individual denied a loan can change their values in order to 
get a positive decision.
 
For each credit card holder in the credit dataset, the label indicates whether a credit card holder will default during the next month ($y=0$) or not ($y=1$) and the features $\xb$ are:
 \begin{itemize}
    \item Marital status: Whether the person is married or single.
    \item Age Group: Group depending on the person's age (<25, 25-39, 40-59, >60).
    \item Education Level: The level of education the individual has acquired (1-4).
    \item Maximum Bill Amount Over Last 6 Months
    \item Maximum Payment Amount Over Last 6 Months
    \item Months With Zero Balance Over Last 6 Months
    \item Months With Low Spending Over Last 6 Months
    \item Months With High Spending Over Last 6 Months
    \item Most Recent Bill Amount
    \item Most Recent Payment Amount
    \item Total Overdue Counts
    \item Total Months Overdue
 \end{itemize}
Here, we assume that all features except Marital Status, Age Group and Education Level are actionable and, among the actionable features, 
we assume that Total Overdue Counts and Total Months Overdue can only increase.

In both cases, note that the actionable features are numerical, however, our methodology only allows for discrete valued features. Therefore, 
rather than using the numerical values as features, we first cluster the loan applicants (or credit card holders) into $k$ groups based on the 
original numerical features using k-clustering and then, for each applicant (or credit card holder), use the cluster identifier it belongs to, 
represented using a one-hot encoding, as a feature.
%
%
%
After this preprocessing step, the discrete feature values $\xb_i$ consists of all possible value combinations of discrete non-actionable features, 
if any, and cluster identifiers.

To approximate the values of the conditional distribution $P(y\given\xb)$, we train four types of classifiers (Multi-layer perceptron, support 
vector machine, logistic regression, decision tree) using the default scikit-learn parameters and then choose the pair of classifier type and 
number of clusters $k$ that maximizes accuracy, estimated using $5$-fold cross validation. Finally, we set $\gamma$ equal to the $50$-th 
percentile of all the individuals' $P(y=1\given\xb)$ values causing a $50\%$ acceptance rate by the optimal threshold policy in the non strategic setting.
Table~\ref{tab:data} summarizes the resulting experimental setup for both datasets.

\begin{table}[t]
\caption{Dataset details }
\begin{center}
\begin{tabular}{ |l||c|c|c|c|c|c|c| } 
 \hline
 Dataset & \# of samples & Classifier & $k$ & Accuracy & $m$ & $\gamma$ \\ \hline 
 credit & $30000$ & Logistic Regression & $100$ & $80.4\%$ & $3200$ & $0.85$ \\ 
 lending & $1266817$ & Logistic Regression & $400$ & $89.9\%$ & $400$ & $0.97$ \\ 
 \hline
\end{tabular}
\end{center}
\end{table}\label{tab:data}

\vspace{-1mm}
\subsection{Examples of counterfactual explanations}
\label{app:realistic}
\vspace{-1mm}

In this section, we focus on the credit dataset and look more closely into the counterfactual explanations $\Ecal_m(\xb)$ and $\Ecal(\xb)$ provided by the minimum cost baseline and 
Algorithm~\ref{alg:greedy}, respectively, by means of an (anecdotal) example.
%
%
%
To this end, for a fixed $\alpha$ and $k$, we first track down the individuals whose best-response under both methods is to change their initial features 
to the provided counterfactual explanation. Then, for each of these individuals, we compare the counterfactual explanations provided by each of both 
methods.
%

Table~\ref{tab:example} shows the initial features $\xb$ together with the counterfactual explanations $\Ecal_m(\xb)$ and $\Ecal(\xb)$ for one of the above 
individuals picked at random.
In this example, the individual is a university student, unmarried and under the age of 25 who is advised to follow the counterfactual explanations to maintain 
her credit.
Since the marital status, age group and level of education are all non-actionable features, both counterfactual explanations maintain the initial values for 
those features.
%
%
%
Under the minimum cost baseline, the bank would advise the individual to reduce her monthly credit card bill by $\sim$\$$150$ and limit high spending to $2$ months 
per semester so that her risk of default would decrease from $16$\% to $13$\%.
However, under Algorithm~\ref{alg:greedy}, the bank would advise to reduce her monthly credit card bill by $\sim$\$$400$, limit high spending to $1$ month per 
semester, and additionally increase her monthly credit card payoff slightly so that her risk of default would decrease to $11\%$.
Since by construction, both $\Ecal_m(\xb)$ and $\Ecal(\xb)$ are inside the region of adaptation of $\xb$, the individual is guaranteed to follow the advice in both cases,
however, under Algorithm~\ref{alg:greedy}, the individual would be less likely to default and achieve a superior long-term well being.
%

%

\begin{table}[t]
\caption{Counterfactual explanations $\Ecal_m(\xb)$ and $\Ecal(\xb)$ provided by the minimum cost baseline and Algorithm~\ref{alg:greedy}, respectively, to an individual 
with initial feature value $\xb$. 
Initially, the individual'{}s outcome is $P(y=1\given\xb)=0.84$ and, after best-response, her outcome is $P(y=1\given\Ecal_m(\xb))=0.87$ and $P(y=1\given\Ecal(\xb))=0.89$,
respectively.
In both methods, we set $\alpha=2$ and $k=160$.
}
\begin{center}
\begin{tabular}{ |l||c|c|c|} 
 \hline
 Feature & $\xb$ & $\Ecal_m(\xb)$ & $\Ecal(\xb)$ \\ \hline 
 Married & No & No & No \\ 
 Age group & Under 25 & Under 25 & Under 25 \\ 
  Education & Student & Student & Student  \\ 
  Maximum Bill Amount Over Last 6 Months & \$$2246$ & \$$2084$ & \$$1929$ \\
  Maximum Payment Amount Over Last 6 Months & \$$191$ & \$$188$ & \$$221$ \\
  Months With Zero Balance Over Last 6 Months & $0$ & $0$ & $0$ \\
  Months With Low Spending Over Last 6 Months & $0$ & $0$ & $0$ \\
  Months With High Spending Over Last 6 Months & $4$ & $2$ & $1$ \\
  Most Recent Bill Amount & \$$2145$ & \$$2003$ & \$$1750$ \\
  Most Recent Payment Amount & \$$123$ & \$$124$ & \$$100$ \\
  Total Overdue Counts & $0$ & $0$ & $0$ \\
  Total Months Overdue & $0$ & $0$ & $0$ \\
 \hline
\end{tabular}
\end{center}
\end{table}\label{tab:example}

%% file: strategic-decisions-examples.bbl
\begin{thebibliography}{48}
\providecommand{\natexlab}[1]{#1}
\providecommand{\url}[1]{\texttt{#1}}
\expandafter\ifx\csname urlstyle\endcsname\relax
  \providecommand{\doi}[1]{doi: #1}\else
  \providecommand{\doi}{doi: \begingroup \urlstyle{rm}\Url}\fi

\bibitem[css()]{css}
Credit score simulator.
\newblock https://www.creditkarma.com/tools/credit-score-simulator/.

\bibitem[len()]{len}
Lending club dataset.
\newblock https://www.kaggle.com/wordsforthewise/lending-club/version/3.

\bibitem[Barocas et~al.(2020)Barocas, Selbst, and Raghavan]{barocas2020hidden}
Solon Barocas, Andrew~D Selbst, and Manish Raghavan.
\newblock The hidden assumptions behind counterfactual explanations and
  principal reasons.
\newblock In \emph{Proceedings of the 2020 Conference on Fairness,
  Accountability, and Transparency}, pages 80--89, 2020.

\bibitem[Br{\"u}ckner and Scheffer(2011)]{bruckner2011stackelberg}
Michael Br{\"u}ckner and Tobias Scheffer.
\newblock Stackelberg games for adversarial prediction problems.
\newblock In \emph{Proceedings of the 17th ACM SIGKDD international conference
  on Knowledge discovery and data mining}, pages 547--555, 2011.

\bibitem[Buchbinder et~al.(2014)Buchbinder, Feldman, Naor, and
  Schwartz]{buchbinder2014submodular}
Niv Buchbinder, Moran Feldman, Joseph Naor, and Roy Schwartz.
\newblock Submodular maximization with cardinality constraints.
\newblock In \emph{Proceedings of the twenty-fifth annual ACM-SIAM symposium on
  Discrete algorithms}, pages 1433--1452. SIAM, 2014.

\bibitem[Calinescu et~al.(2011)Calinescu, Chekuri, Pal, and
  Vondr{\'a}k]{calinescu2011maximizing}
Gruia Calinescu, Chandra Chekuri, Martin Pal, and Jan Vondr{\'a}k.
\newblock Maximizing a monotone submodular function subject to a matroid
  constraint.
\newblock \emph{SIAM Journal on Computing}, 40\penalty0 (6):\penalty0
  1740--1766, 2011.

\bibitem[Chakraborty et~al.(2017)Chakraborty, Tomsett, Raghavendra, Harborne,
  Alzantot, Cerutti, Srivastava, Preece, Julier, Rao,
  et~al.]{chakraborty2017interpretability}
Supriyo Chakraborty, Richard Tomsett, Ramya Raghavendra, Daniel Harborne,
  Moustafa Alzantot, Federico Cerutti, Mani Srivastava, Alun Preece, Simon
  Julier, Raghuveer~M Rao, et~al.
\newblock Interpretability of deep learning models: a survey of results.
\newblock In \emph{2017 IEEE SmartWorld, Ubiquitous Intelligence \& Computing,
  Advanced \& Trusted Computed, Scalable Computing \& Communications, Cloud \&
  Big Data Computing, Internet of People and Smart City Innovation
  (SmartWorld/SCALCOM/UIC/ATC/CBDCom/IOP/SCI)}, pages 1--6. IEEE, 2017.

\bibitem[Coate and Loury(1993)]{coate1993will}
S.~Coate and G.~Loury.
\newblock Will affirmative-action policies eliminate negative stereotypes?
\newblock \emph{The American Economic Review}, 1993.

\bibitem[Corbett-Davies et~al.(2017)Corbett-Davies, Pierson, Feller, Goel, and
  Huq]{corbett2017algorithmic}
Sam Corbett-Davies, Emma Pierson, Avi Feller, Sharad Goel, and Aziz Huq.
\newblock Algorithmic decision making and the cost of fairness.
\newblock In \emph{Proceedings of the 23rd ACM SIGKDD International Conference
  on Knowledge Discovery and Data Mining}, pages 797--806, 2017.

\bibitem[Dalvi et~al.(2004)Dalvi, Domingos, Sanghai, and
  Verma]{dalvi2004adversarial}
Nilesh Dalvi, Pedro Domingos, Sumit Sanghai, and Deepak Verma.
\newblock Adversarial classification.
\newblock In \emph{Proceedings of the tenth ACM SIGKDD international conference
  on Knowledge discovery and data mining}, pages 99--108, 2004.

\bibitem[Dong et~al.(2018)Dong, Roth, Schutzman, Waggoner, and
  Wu]{dong2018strategic}
Jinshuo Dong, Aaron Roth, Zachary Schutzman, Bo~Waggoner, and Zhiwei~Steven Wu.
\newblock Strategic classification from revealed preferences.
\newblock In \emph{Proceedings of the 2018 ACM Conference on Economics and
  Computation}, pages 55--70, 2018.

\bibitem[Doshi-Velez and Kim(2017)]{doshi2017towards}
Finale Doshi-Velez and Been Kim.
\newblock Towards a rigorous science of interpretable machine learning.
\newblock \emph{arXiv preprint arXiv:1702.08608}, 2017.

\bibitem[Fryer and Loury(2013)]{fryer2013valuing}
R.~Fryer and G.~Loury.
\newblock Valuing diversity.
\newblock \emph{Journal of Political Economy}, 2013.

\bibitem[Gunning and Aha(2019)]{gunning2019darpa}
David Gunning and David~W Aha.
\newblock Darpa's explainable artificial intelligence program.
\newblock \emph{AI Magazine}, 40\penalty0 (2):\penalty0 44--58, 2019.

\bibitem[Hardt et~al.(2016{\natexlab{a}})Hardt, Megiddo, Papadimitriou, and
  Wootters]{hardt2016strategic}
Moritz Hardt, Nimrod Megiddo, Christos Papadimitriou, and Mary Wootters.
\newblock Strategic classification.
\newblock In \emph{Proceedings of the 2016 ACM conference on innovations in
  theoretical computer science}, pages 111--122, 2016{\natexlab{a}}.

\bibitem[Hardt et~al.(2016{\natexlab{b}})Hardt, Price, and
  Srebro]{hardt2016equality}
Moritz Hardt, Eric Price, and Nati Srebro.
\newblock Equality of opportunity in supervised learning.
\newblock In \emph{Advances in neural information processing systems}, pages
  3315--3323, 2016{\natexlab{b}}.

\bibitem[Hochbaum and Pathria(1998)]{hochbaum1998analysis}
Dorit~S Hochbaum and Anu Pathria.
\newblock Analysis of the greedy approach in problems of maximum k-coverage.
\newblock \emph{Naval Research Logistics (NRL)}, 45\penalty0 (6):\penalty0
  615--627, 1998.

\bibitem[Hu and Chen(2018)]{hu2018short}
L.~Hu and Y.~Chen.
\newblock A short-term intervention for long-term fairness in the labor market.
\newblock In \emph{WWW}, 2018.

\bibitem[Hu et~al.(2019)Hu, Immorlica, and Vaughan]{hu2019disparate}
Lily Hu, Nicole Immorlica, and Jennifer~Wortman Vaughan.
\newblock The disparate effects of strategic manipulation.
\newblock In \emph{Proceedings of the Conference on Fairness, Accountability,
  and Transparency}, pages 259--268, 2019.

\bibitem[Karimi et~al.(2019)Karimi, Barthe, Belle, and Valera]{karimi2019model}
Amir-Hossein Karimi, Gilles Barthe, Borja Belle, and Isabel Valera.
\newblock Model-agnostic counterfactual explanations for consequential
  decisions.
\newblock \emph{arXiv preprint arXiv:1905.11190}, 2019.

\bibitem[Karp(1972)]{karp1972reducibility}
Richard~M Karp.
\newblock Reducibility among combinatorial problems.
\newblock In \emph{Complexity of computer computations}, pages 85--103.
  Springer, 1972.

\bibitem[Kilbertus et~al.(2019)Kilbertus, Gomez-Rodriguez, Sch\"{o}lkopf,
  Muandet, and Valera]{kilbertus2019fair}
Niki Kilbertus, Manuel Gomez-Rodriguez, Bernhard Sch\"{o}lkopf, Krikamol
  Muandet, and Isabel Valera.
\newblock Fair decisions despite imperfect predictions.
\newblock In \emph{AISTATS}, 2019.

\bibitem[Kleinberg and Raghavan(2019)]{kleinberg2019classifiers}
Jon Kleinberg and Manish Raghavan.
\newblock How do classifiers induce agents to invest effort strategically?
\newblock In \emph{Proceedings of the 2019 ACM Conference on Economics and
  Computation}, pages 825--844, 2019.

\bibitem[Kleinberg et~al.(2018)Kleinberg, Lakkaraju, Leskovec, Ludwig, and
  Mullainathan]{kleinberg2018human}
Jon Kleinberg, Himabindu Lakkaraju, Jure Leskovec, Jens Ludwig, and Sendhil
  Mullainathan.
\newblock Human decisions and machine predictions.
\newblock \emph{The quarterly journal of economics}, 133\penalty0 (1):\penalty0
  237--293, 2018.

\bibitem[Koh and Liang(2017)]{koh2017understanding}
Pang~Wei Koh and Percy Liang.
\newblock Understanding black-box predictions via influence functions.
\newblock In \emph{Proceedings of the 34th International Conference on Machine
  Learning}, 2017.

\bibitem[Lipton(2018)]{lipton2018mythos}
Zachary~C Lipton.
\newblock The mythos of model interpretability.
\newblock \emph{Queue}, 16\penalty0 (3):\penalty0 31--57, 2018.

\bibitem[Liu et~al.(2018)Liu, Dean, Rolf, Simchowitz, and
  Hardt]{liu2018delayed}
Lydia~T Liu, Sarah Dean, Esther Rolf, Max Simchowitz, and Moritz Hardt.
\newblock Delayed impact of fair machine learning.
\newblock In \emph{Advances in neural information processing systems}, 2018.

\bibitem[Lundberg and Lee(2017)]{lundberg2017unified}
Scott~M Lundberg and Su-In Lee.
\newblock A unified approach to interpreting model predictions.
\newblock In \emph{Advances in neural information processing systems}, 2017.

\bibitem[Miller et~al.(2019)Miller, Milli, and Hardt]{miller2019strategic}
John Miller, Smitha Milli, and Moritz Hardt.
\newblock Strategic adaptation to classifiers: A causal perspective.
\newblock \emph{arXiv preprint arXiv:1910.10362}, 2019.

\bibitem[Milli et~al.(2019)Milli, Miller, Dragan, and Hardt]{milli2019social}
Smitha Milli, John Miller, Anca~D Dragan, and Moritz Hardt.
\newblock The social cost of strategic classification.
\newblock In \emph{Proceedings of the Conference on Fairness, Accountability,
  and Transparency}, pages 230--239, 2019.

\bibitem[Mitchell et~al.(2018)Mitchell, Potash, Barocas, D'Amour, and
  Lum]{mitchell2018prediction}
Shira Mitchell, Eric Potash, Solon Barocas, Alexander D'Amour, and Kristian
  Lum.
\newblock Prediction-based decisions and fairness: A catalogue of choices,
  assumptions, and definitions.
\newblock \emph{arXiv preprint arXiv:1811.07867}, 2018.

\bibitem[Mothilal et~al.(2020)Mothilal, Sharma, and
  Tan]{mothilal2019explaining}
Ramaravind~K. Mothilal, Amit Sharma, and Chenhao Tan.
\newblock Explaining machine learning classifiers through diverse
  counterfactual explanations.
\newblock In \emph{Proceedings of the 2020 Conference on Fairness,
  Accountability, and Transparency}, 2020.

\bibitem[Murdoch et~al.(2019)Murdoch, Singh, Kumbier, Abbasi-Asl, and
  Yu]{murdoch2019definitions}
W~James Murdoch, Chandan Singh, Karl Kumbier, Reza Abbasi-Asl, and Bin Yu.
\newblock Definitions, methods, and applications in interpretable machine
  learning.
\newblock \emph{Proceedings of the National Academy of Sciences}, 116\penalty0
  (44):\penalty0 22071--22080, 2019.

\bibitem[Nemhauser et~al.(1978)Nemhauser, Wolsey, and
  Fisher]{nemhauser1978analysis}
George~L Nemhauser, Laurence~A Wolsey, and Marshall~L Fisher.
\newblock An analysis of approximations for maximizing submodular set
  functions—i.
\newblock \emph{Mathematical programming}, 14\penalty0 (1):\penalty0 265--294,
  1978.

\bibitem[Perdomo et~al.(2020)Perdomo, Zrnic, Mendler-D{\"u}nner, and
  Hardt]{perdomo2020performative}
Juan~C Perdomo, Tijana Zrnic, Celestine Mendler-D{\"u}nner, and Moritz Hardt.
\newblock Performative prediction.
\newblock \emph{arXiv preprint arXiv:2002.06673}, 2020.

\bibitem[Ribeiro et~al.(2016)Ribeiro, Singh, and Guestrin]{ribeiro2016should}
Marco~Tulio Ribeiro, Sameer Singh, and Carlos Guestrin.
\newblock Why should i trust you? explaining the predictions of any classifier.
\newblock In \emph{Proceedings of the 22nd ACM SIGKDD international conference
  on knowledge discovery and data mining}, 2016.

\bibitem[Rudin(2019)]{rudin2019stop}
Cynthia Rudin.
\newblock Stop explaining black box machine learning models for high stakes
  decisions and use interpretable models instead.
\newblock \emph{Nature Machine Intelligence}, 1\penalty0 (5):\penalty0
  206--215, 2019.

\bibitem[Russell(2019)]{russell2019efficient}
Chris Russell.
\newblock Efficient search for diverse coherent explanations.
\newblock In \emph{Proceedings of the Conference on Fairness, Accountability,
  and Transparency}, pages 20--28, 2019.

\bibitem[Solis-Oba(2006)]{solis2006approximation}
Roberto Solis-Oba.
\newblock Approximation algorithms for the k-median problem.
\newblock In \emph{Efficient Approximation and Online Algorithms}, pages
  292--320. Springer, 2006.

\bibitem[Tabibian et~al.(2020)Tabibian, Tsirtsis, Khajehnejad, Singla,
  Sch\"{o}lkopf, and Gomez-Rodriguez]{tabibian2020optimal}
Behzad Tabibian, Stratis Tsirtsis, Moein Khajehnejad, Adish Singla, Bernhard
  Sch\"{o}lkopf, and Manuel Gomez-Rodriguez.
\newblock Optimal decision making under strategic behavior.
\newblock \emph{Arxiv:1905.09239}, 2020.

\bibitem[Tolomei et~al.(2017)Tolomei, Silvestri, Haines, and
  Lalmas]{tolomei2017interpretable}
Gabriele Tolomei, Fabrizio Silvestri, Andrew Haines, and Mounia Lalmas.
\newblock Interpretable predictions of tree-based ensembles via actionable
  feature tweaking.
\newblock In \emph{Proceedings of the 23rd ACM SIGKDD international conference
  on knowledge discovery and data mining}, pages 465--474, 2017.

\bibitem[Ustun et~al.(2019)Ustun, Spangher, and Liu]{ustun2019actionable}
Berk Ustun, Alexander Spangher, and Yang Liu.
\newblock Actionable recourse in linear classification.
\newblock In \emph{Proceedings of the Conference on Fairness, Accountability,
  and Transparency}, pages 10--19, 2019.

\bibitem[Valera et~al.(2018)Valera, Singla, and Rodriguez]{valera2018enhancing}
Isabel Valera, Adish Singla, and Manuel~Gomez Rodriguez.
\newblock Enhancing the accuracy and fairness of human decision making.
\newblock In \emph{Advances in Neural Information Processing Systems}, pages
  1769--1778, 2018.

\bibitem[Voigt and Von~dem Bussche(2017)]{voigt2017eu}
Paul Voigt and Axel Von~dem Bussche.
\newblock The eu general data protection regulation (gdpr).
\newblock \emph{A Practical Guide, 1st Ed., Cham: Springer International
  Publishing}, 2017.

\bibitem[Wachter et~al.(2017{\natexlab{a}})Wachter, Mittelstadt, and
  Floridi]{wachter2017right}
Sandra Wachter, Brent Mittelstadt, and Luciano Floridi.
\newblock Why a right to explanation of automated decision-making does not
  exist in the general data protection regulation.
\newblock \emph{International Data Privacy Law}, 7\penalty0 (2):\penalty0
  76--99, 2017{\natexlab{a}}.

\bibitem[Wachter et~al.(2017{\natexlab{b}})Wachter, Mittelstadt, and
  Russell]{wachter2017counterfactual}
Sandra Wachter, Brent Mittelstadt, and Chris Russell.
\newblock Counterfactual explanations without opening the black box: Automated
  decisions and the gdpr.
\newblock \emph{Harv. JL \& Tech.}, 31:\penalty0 841, 2017{\natexlab{b}}.

\bibitem[Weller(2017)]{weller2017challenges}
Adrian Weller.
\newblock Challenges for transparency.
\newblock 2017.

\bibitem[Yeh and Lien(2009)]{yeh2009comparisons}
I-Cheng Yeh and Che-hui Lien.
\newblock The comparisons of data mining techniques for the predictive accuracy
  of probability of default of credit card clients.
\newblock \emph{Expert Systems with Applications}, 36\penalty0 (2):\penalty0
  2473--2480, 2009.

\end{thebibliography}
